\pdfoutput=1

\documentclass[11pt]{article}

\usepackage[preprint]{acl}

\usepackage{times}
\usepackage{latexsym}

\usepackage[T1]{fontenc}

\usepackage[utf8]{inputenc}

\usepackage{microtype}

\usepackage{inconsolata}

\usepackage{graphicx}

\usepackage{booktabs}       
\usepackage{amsfonts}       
\usepackage{nicefrac}       
\usepackage{subcaption}     
\usepackage{adjustbox}      
\usepackage{amsmath}        
\usepackage{multirow}       
\usepackage{hhline}         
\usepackage{svg}            
\usepackage{tikz}           
\usepackage{listings}
\usepackage{tcolorbox}

\newcommand{\accord}{\texttt{ACCORD}}

\newcommand{\makeblue}[1]{\textcolor[HTML]{1F77B4}{#1}}
\newcommand{\blueemph}[1]{\makeblue{\textbf{\textit{#1}}}}
\newcommand{\makeorange}[1]{\textcolor[HTML]{FF7F0E}{#1}}
\newcommand{\orangeemph}[1]{\makeorange{\textbf{\textit{#1}}}}
\newcommand{\factual}{\blueemph{factual}}
\newcommand{\Factual}{\blueemph{Factual}}
\newcommand{\antifactual}{\orangeemph{anti-factual}}
\newcommand{\Antifactual}{\orangeemph{Anti-factual}}
\newcommand{\AntiFactual}{\orangeemph{Anti-Factual}}
\newcommand{\wdef}{\makeblue{$\mathbf{w}^\text{def}$}}
\newcommand{\waf}{\makeorange{$\mathbf{w}^\text{af}$}}
\newcommand{\adef}{\makeblue{$a^\text{def}$}}
\newcommand{\aaf}{\makeorange{$a^\text{af}$}}
\newcommand{\caf}{\makeorange{$C^\text{af}$}}

\newcommand{\emphbox}[2]{\colorbox{gray!10}{\begin{minipage}{#1\linewidth}#2\end{minipage}}}
\newcommand{\doubleemphbox}[1]{\colorbox{gray!30}{#1}}

\newcommand*\emptycirc[1][1ex]{\tikz\draw (0,0) circle (#1);} 
\newcommand*\halfcirc[1][1ex]{%
  \begin{tikzpicture}
  \draw[fill] (0,0) -- (90:#1) arc (90:270:#1) -- cycle ;
  \draw (0,0) circle (#1);
  \end{tikzpicture}}
\newcommand*\fullcirc[1][1ex]{\tikz\fill (0,0) circle (#1);}

\newcolumntype{R}[2]{%
    >{\adjustbox{angle=#1,lap=\width-(#2)}\bgroup}%
    l%
    <{\egroup}%
}
\newcommand*\rot{\multicolumn{3}{R{30}{1em}}}

\title{\accord: Closing the Commonsense Measurability Gap}

\author{
  \textbf{Fran{\c{c}}ois Roewer-Despr{\'e}s\textsuperscript{1,4}},
  \textbf{Jinyue Feng\textsuperscript{1,4}},
  \textbf{Zining Zhu\textsuperscript{2,4}},
  \textbf{Frank Rudzicz\textsuperscript{3,4}},
\\
\\
  \textsuperscript{1}University of Toronto,
  \textsuperscript{2}Stevens Institute of Technology,
  \textsuperscript{3}Dalhousie University,
  \textsuperscript{4}Vector Institute
\\
  \small{
    \textbf{Correspondence:} \href{mailto:francoisrd@cs.toronto.edu}{francoisrd@cs.toronto.edu}
  }
}

\begin{document}
\maketitle

\begin{abstract}
We present \accord, a framework and benchmark suite for disentangling the commonsense grounding and reasoning abilities of large language models (LLMs) through controlled, multi-hop counterfactuals. \accord{}
introduces formal elements to
commonsense reasoning to explicitly control and quantify reasoning complexity beyond the typical 1 or 2 hops. Uniquely, \accord{} can automatically generate benchmarks of arbitrary reasoning complexity, so it scales with future LLM improvements. Indeed, our experiments on state-of-the-art LLMs show performance degrading to \textit{below random chance} with only moderate scaling, leaving substantial headroom for improvement. We release a leaderboard\footnote{Leaderboard and dataset download: \url{https://www.codabench.org/competitions/3160/}} of the benchmark suite tested in this work, as well as code\footnote{Source code: \url{https://github.com/francois-rd/accord/}} to automatically generate more complex benchmarks.
\end{abstract}

\section{Introduction}
\label{sec:intro}

\begin{table*}[t]
\centering
\begin{adjustbox}{max width=\linewidth}
\begin{tabular}{p{0.16\linewidth}p{0.25\linewidth}p{0.65\linewidth}}
\toprule
\textbf{Counterfactual} & \textbf{Definition} & \textbf{Example} \\
\midrule
Hypothetical & Runs counter to scenario, but plausible under \wdef & A bird flies over a bridge [\textit{scenario}]. What would have happened if the bird had hit the bridge? \cite{frohberg2021crass} \\
\makeorange{Anti-Factual} & Implausible under \wdef & If cats were vegetarians, cats would love cabbages. \cite{li2023counterfactual} \\
\bottomrule
\end{tabular}
\end{adjustbox}
\caption{Differentiating hypothetical from \antifactual{} counterfactuals. \wdef{} is a model of the default (i.e., commonsense) worldview underpinning the
vast
majority of existing LLM training data \cite{wu2023reasoning}. LLMs trained on such data can parrot answers that spuriously circumvent hypothetical reasoning tasks. \orangeemph{Anti-factuals} are significantly stronger mitigators of this inductive bias precisely because they are implausible under the training data.}
\label{tab:counterfactual}
\end{table*}

\begin{figure*}[t]
\begin{adjustbox}{max width=\linewidth}
\centering
\begin{subfigure}{0.30\linewidth}
\textbf{Rule:}

$\texttt{part\_of}(x,y) \land \texttt{spatial}(y,z)$

\hspace{1em} $\Rightarrow \texttt{spatial}(x,z)$

\textbf{Facts:}

\hspace{0.25em} $\texttt{part\_of}(hydrogen, water)$

\hspace{0.25em} $\texttt{spatial}(water, ocean)$

\textbf{Conclusion:}

\hspace{0.25em} $\texttt{spatial}(hydrogen, ocean)$
\caption{Formal}
\label{fig:formal-vs-commonsense:formal}
\end{subfigure}
\begin{subfigure}{0.01\linewidth}
\hfill
\vfill
\end{subfigure}
\begin{subfigure}{0.35\linewidth}
\textbf{Rule:}

<same rule, but must be implicitly inferred using commonsense>

\textbf{Facts:}

\hspace{0.25em} hydrogen is part of water molecules

\hspace{0.25em} water molecules appear near oceans

\textbf{Conclusion:}

\hspace{0.25em} hydrogen appears near oceans
\caption{Commonsense}
\label{fig:formal-vs-commonsense:commonsense}
\end{subfigure}
\begin{subfigure}{0.01\linewidth}
\hfill
\vfill
\end{subfigure}
\begin{subfigure}{0.30\linewidth}
\textbf{Rule:}

<same rule, but must be inferred using commonsense>

\textbf{Facts:}

\hspace{0.25em} $\texttt{part\_of}(hydrogen, water)$

\hspace{0.25em} $\texttt{spatial}(water, ocean)$

\textbf{Conclusion:}

\hspace{0.25em} $\texttt{spatial}(hydrogen, ocean)$
\caption{Formalized Commonsense}
\label{fig:formal-vs-commonsense:formal-commonsense}
\end{subfigure}
\begin{subfigure}{0.01\linewidth}
\hfill
\vfill
\end{subfigure}
\begin{subfigure}{0.35\linewidth}
\textbf{Rule:}

<same rule, but must be implicitly inferred using commonsense>

\textbf{Facts:}

\hspace{0.25em} $\texttt{part\_of}(outer~space, watch)$

\hspace{0.25em} $\texttt{spatial}(watch, the~planet)$

\textbf{Conclusion:}

\hspace{0.25em} $\texttt{spatial}(outer~space, the~planet)$
\caption{\AntiFactual{} Commonsense}
\label{fig:formal-vs-commonsense:af-commonsense}
\end{subfigure}
\end{adjustbox}
\caption{Differentiating types of reasoning. \textbf{(a)} Formal reasoning requires systematically applying formal rules, \textit{which must be provided explicitly} \cite{huang2022towards}. In this example, the formal rule states that, for any objects $x$, $y$, and $z$, if $x$ is a component part of (and therefore near) $y$ and $y$ is near $z$, then $x$ is near $z$. Notice that the conclusion cannot be \textit{formally} derived from the facts without this rule. \textbf{(b)} Commonsense relies on prior knowledge (the semantics of ``part of'' and ``spatial'') to implicitly fill in knowledge gaps (the omitted rule), which is ill-defined \cite{huang2022towards, davis2023benchmarks}. \textbf{(c)} Formalized commonsense (\textbf{our work}) formalizes the reasoning elements while maintaining implicit knowledge gaps, which is well-defined and enables the automated verification of correctness. \textbf{(d)} \Antifactual{} formalized commonsense (\textbf{our work}) maintains the same formal reasoning elements and underlying rule, but grounds variables $x$, $y$, and $z$ with implausible objects to prevent LLMs from spuriously parroting the conclusion without having first reasoned through the facts \cite{wu2023reasoning}.}
\label{fig:formal-vs-commonsense}
\end{figure*}

LLMs perform remarkably well on diverse reasoning tasks \cite{wei2022emergent, wei2022chain, kojima2022large}. However, more detailed analysis reveals their reasoning remains unreliable \cite{huang2022towards, valmeekam2022large, bang2023multitask, yang2023coupling}. LLM reasoning often lacks robustness to simple lexical triggers \cite{li2023counterfactual, pandia2021sorting} or irrelevant context \cite{misra2022comps, shi2023large}. In Chain-of-Thought (CoT) approaches, LLMs can decompose problems into illogical or irrelevant reasoning chains \cite{wei2022chain, kojima2022large, prasad2023receval, wang2023scott} and can systematically rationalize incorrect conclusions based on a simple reordering of answer choices \cite{turpin2024language}. Furthermore, these limitations appear to worsen as the complexity of a reasoning task increases \cite{xu2023large, dziri2024faith}.

Separately, LLMs are increasingly being used as knowledge bases (KBs) \cite{petroni2019language, wu2023chain} given their impressive ability to store knowledge in model parameters. However, such parametric knowledge incurs limitations \cite{welleck2019neural, baek2023knowledge, ji2023survey}. For example, \citet{zhou2023context} report that, given the context \textit{``Elon Musk [...] is the owner and CEO of Twitter''}, GPT-3.5 answered the question \textit{``Who is the CEO of Twitter?''} with \textit{``Jack Dorsey''} (the previous CEO), since Musk became CEO only after GPT-3.5's September 2021 knowledge cutoff.

Grounding of LLMs with contextual knowledge during inference has become a leading paradigm (e.g., retrieval-augmented generation). However, as the above example illustrates, although their parametric knowledge can become outdated \cite{cheng2023decouple, onoe2023can, xu2023kilm, gao2023retrieval, zheng2023can}, LLMs tend to integrate contextual knowledge unfaithfully \cite{arodi2022kitmus, huang2023zero, ji2023rho, sun2023towards}, especially when parametric and contextual knowledge directly conflict \cite{longpre2021entity, li2022large, neeman2022disentqa, li2023counterfactual, monea2023glitch, tang2023large, yu2023characterizing}.

\begin{table*}[t]
\centering
\begin{adjustbox}{max width=\linewidth}
\begin{tabular}{ll|lll}
\toprule
\textbf{Reasoning Skill} & \textbf{Definition} & \textbf{ConceptNet} & \textbf{Reasoning Template} & \textbf{Name} \\
\midrule
Spatial         & X appears near Y                      & \texttt{AtLocation}      & $\mathbf{X}$ appears near $\mathbf{Y}$ & \texttt{spatial} \\
Cause \& Effect & X causes Y                            & \texttt{Causes}          & $\mathbf{X}$ causes $\mathbf{Y}$ & \texttt{causal} \\
Has Parts       & X contains Y as one of its parts      & \texttt{PartOf}          & $\mathbf{Y}$ is a part of $\mathbf{X}$ & \texttt{part\_of} \\
Is Member Of    & X belongs to the larger class of Y    & \texttt{IsA}             & $\mathbf{X}$ is a type of $\mathbf{Y}$ & \texttt{type\_of} \\
Purpose         & X is the purpose of Y                 & \texttt{UsedFor}         & $\mathbf{Y}$ is used for $\mathbf{X}$ & \texttt{used\_for} \\
Preconditions   & X must hold true for Y to take place  & \texttt{HasPrerequisite} & $\mathbf{Y}$ has prerequisite $\mathbf{X}$ & \texttt{requires} \\
\bottomrule
\end{tabular}
\end{adjustbox}
\caption{Subset of reasoning skills and definitions identified by \citealt{talmor2018commonsenseqa}, along with a mapping to an appropriate ConceptNet \cite{speer2017conceptnet} relation and associated named reasoning template (\textbf{our work}).}
\label{table:reasoning-skills-and-templates}
\end{table*}

Unfortunately, this so-called `context unfaithfulness' confounds empirical measurements of reasoning, leading to construct validity concerns \cite{mccoy2019right, kiciman2023causal, prasad2023receval}. Indeed, LLMs can circumvent the intended objective of various reasoning tasks using inductive biases from parametric knowledge \cite{longpre2021entity, dziri2024faith}, dataset artifacts \cite{ho2020constructing, turpin2024language}, and other spurious shortcuts \cite{feng2022generic, guo2022counterfactual, chen2023say, yuan2023causality}. More formally, borrowing from \citet{wu2023reasoning}, we conceptualize a \blueemph{default} world model, \wdef, as the set of conditions and assumptions underpinning the vast majority of existing LLM training data.
Under this view, LLMs can spuriously circumvent \wdef-grounded reasoning tasks precisely because they are trained on \wdef{} data \cite{wu2023reasoning}.

\vspace{-0.5em}
Fortunately, counterfactual grounding, in which (at least some) contextual knowledge runs counter to \wdef, can mitigate this concern \cite{chan2023spurious, li2023counterfactual}. However, the effectiveness of counterfactual grounding depends on the degree to which LLMs cease resorting to parametric shortcuts. To illustrate this point, we distinguish between two types of counterfactuals (see Table \ref{tab:counterfactual}): \textbf{\textit{hypothetical}}, in which the context runs counter to a specific scenario, but remains plausible under \wdef; and \antifactual{} (AF), in which the context is implausible under \wdef. LLMs are less likely to spuriously solve \orangeemph{anti-factuals} compared to hypotheticals precisely because the former are implausible under \wdef{} and are thus significantly stronger inductive bias mitigators \cite{wu2023reasoning}.

\vspace{-0.5em}
Reasoning is generally categorized into subtypes \cite{huang2022towards, mialon2023augmented, qiao2022reasoning}, most of which---including mathematical, logical, and symbolic reasoning---are types of formal reasoning that follow systematic rules and logical processes \cite{huang2022towards}. All knowledge needed to solve formal reasoning problems \textit{must be provided} as explicit rules (see Figure \ref{fig:formal-vs-commonsense:formal}). In contrast, commonsense reasoning relies on informal reasoning to fill in knowledge gaps based on intuition and worldly experience (see Figure \ref{fig:formal-vs-commonsense:commonsense}) \cite{huang2022towards, davis2023benchmarks}. This makes commonsense reasoning well-suited to our goal, since LLMs learn this worldly experience via pretraining on \wdef, which is prone to grounding confounds \cite{wu2023reasoning}. Commonsense reasoning has a broad scope, and its boundaries are vague and ill-defined \cite{davis2023benchmarks}. As a result, it can generally be understood as a catch-all term for all non-formal reasoning.

\vspace{-0.25em}
However, ``non-formal'' does not imply a lack of reasoning complexity \cite{davis2023benchmarks}. Indeed, commonsense reasoning tasks aim to test precisely this ability. Unfortunately, as established above, \blueemph{factually}-grounded tasks cannot distinguish between genuine reasoning ability and mere parametric parroting of a suitable common knowledge solution. Being able to explicitly control and quantify reasoning complexity is a feature ubiquitous in formal reasoning benchmarks, but overwhelmingly lacking in commonsense benchmarks \cite{davis2023benchmarks}.

We present \accord, a framework for generating \textbf{A}nti-fa\textbf{C}tual \textbf{CO}mmonsense \textbf{R}easoning \textbf{D}isentanglement benchmarks. \accord{} tightly controls fine-grained \antifactual{} variants (see Figure \ref{fig:formal-vs-commonsense:af-commonsense}) of commonsense reasoning tasks to enable detailed analysis of LLM performance factors. Through \orangeemph{anti-factuals}, \accord{} disentangles commonsense (contextual) grounding and (parametric) reasoning abilities in LLMs. Since the lack of formal rules renders controlled analysis of commonsense reasoning particularly challenging, \accord{} borrows from formal reasoning to partially formalize commonsense reasoning (see Figure \ref{fig:formal-vs-commonsense:formal-commonsense}), while explicitly controlling and quantifying reasoning complexity. In this way, \accord{} takes a significant step towards closing the commonsense measurability gap with respect to formal reasoning.
Moreover, \accord{} is uniquely designed to generate future benchmarks of arbitrary reasoning complexity with minimal additional human effort, and, as such, automatically scales its difficulty level in tandem with future improvements in LLM abilities.

\section{\accord{} Framework}
\label{sec:accord} 

\begin{figure*}[t]
\centering
\begin{minipage}{0.48\linewidth}
\begin{subfigure}[t]{\linewidth}
\begin{small}
\emphbox{0.99}{
\textbf{Question:}

Where is the first place someone leaving the planet ends up?

\textbf{Answer choices:}

A: pay debts\hspace{1em}B: galaxy\hspace{1em}C: \blueemph{outer space}\hspace{1em}D: orbit
}
\end{small}
\caption{Original CSQA Instance}
\label{fig:csqa:original}
\end{subfigure}
\end{minipage}
\begin{minipage}{0.02\linewidth}
\hfill
\end{minipage}
\begin{minipage}{0.48\linewidth}
\begin{subfigure}[t]{\linewidth}
\begin{small}
\emphbox{0.8}{
\textbf{Pairing Template:}

-- Suppose that [$\mathbf{V}_x$] does \{not\} appear near [$\mathbf{V}_p$]

\textbf{Pairing Term:}

$\mathbf{V}_p \leftarrow$ \texttt{the planet}
}
\end{small}
\caption{Answer-Discriminating Pairing Template}
\label{fig:csqa:template}
\end{subfigure}
\end{minipage}
%
%
%
%
%
%
\begin{minipage}{0.48\linewidth}
\begin{subfigure}[t]{\linewidth}
\begin{small}
\vspace{1em}
\emphbox{0.65}{
\textbf{Example of Applied Filters:}

\hspace{2em}\texttt{size=2}

\hspace{2em}\texttt{pairing=``appears near''}

\textbf{Example of Match:}

-- Suppose that [$\mathbf{V}_1$] is a part of [$\mathbf{V}_2$]

-- Suppose that [$\mathbf{V}_2$] appears near [$\mathbf{V}_3$]
}
\end{small}
\caption{Generic Reasoning Tree Matching Pairing Template}
\label{fig:csqa:generic_tree}
\end{subfigure}
\begin{subfigure}[t]{\linewidth}
\begin{small}
\vspace{1em}
\emphbox{0.95}{
-- Suppose that [\doubleemphbox{$\mathbf{V}_a$}] is a part of [$\mathbf{V}_2$]

-- Suppose that [$\mathbf{V}_2$] \doubleemphbox{does \{not\}} appear near [\doubleemphbox{the planet}]
}
\end{small}
\caption{Paired Reasoning Tree with 2-hop Reasoning Path}
\label{fig:csqa:paired_tree}
\end{subfigure}
\end{minipage}
\begin{minipage}{0.02\linewidth}
\hfill
\end{minipage}
\begin{minipage}{0.48\linewidth}
\begin{subfigure}[t]{\linewidth}
\begin{small}
\vspace{1em}
\emphbox{0.93}{
-- Suppose that [\doubleemphbox{pay debts}] is a part of [\doubleemphbox{$\mathbf{V}_A$}]

-- Suppose that [\doubleemphbox{$\mathbf{V}_A$}] does \{not\} appear near [the planet]

\vspace{0.5em}

-- Suppose that [\doubleemphbox{galaxy}] is a part of [\doubleemphbox{$\mathbf{V}_B$}]

-- Suppose that [\doubleemphbox{$\mathbf{V}_D$}] does \{not\} appear near [the planet]

\vspace{0.5em}

-- Suppose that [\doubleemphbox{outer space}] is a part of [\doubleemphbox{$\mathbf{V}_C$}]

-- Suppose that [\doubleemphbox{$\mathbf{V}_C$}] does \{not\} appear near [the planet]

\vspace{0.5em}

-- Suppose that [\doubleemphbox{orbit}] is a part of [\doubleemphbox{$\mathbf{V}_D$}]

-- Suppose that [\doubleemphbox{$\mathbf{V}_D$}] does \{not\} appear near [the planet]
}
\end{small}
\caption{Tree Duplication to Avoid Lexical Matching Bias}
\label{fig:csqa:duplication}
\end{subfigure}
\end{minipage}
%
%
%
%
%
%
\begin{minipage}{0.48\linewidth}
\begin{subfigure}[t]{\linewidth}
\begin{small}
\vspace{1em}
\emphbox{0.99}{
\doubleemphbox{\textbf{Statements:}}

-- Suppose that [pay debts] is a part of [\doubleemphbox{forest}]

-- Suppose that [\doubleemphbox{forest}] does \doubleemphbox{\textcolor[HTML]{E06666}{\textbf{not}}} appear near [the planet]

\vspace{0.5em}

-- Suppose that [galaxy] is a part of [\doubleemphbox{coat}]

-- Suppose that [\doubleemphbox{coat}] does \doubleemphbox{\textcolor[HTML]{E06666}{\textbf{not}}} appear near [the planet]

\vspace{0.5em}

-- Suppose that [outer space] is a part of [\doubleemphbox{watch}]

-- Suppose that [\doubleemphbox{watch}] \doubleemphbox{\blueemph{does}} appear near [the planet]

\vspace{0.5em}

-- Suppose that [orbit] is a part of [\doubleemphbox{story}]

-- Suppose that [\doubleemphbox{story}] does \doubleemphbox{\textcolor[HTML]{E06666}{\textbf{not}}} appear near [the planet]

\textbf{Question:}

Where is the first place someone leaving the planet ends up?

\textbf{Answer choices:}

A: pay debts\hspace{1em}B: galaxy\hspace{1em}C: \doubleemphbox{\blueemph{outer space}}\hspace{1em}D: orbit
}
\end{small}
\caption{CSQA Augmented with \doubleemphbox{\Factual{}} Reasoning}
\label{fig:csqa:factual}
\end{subfigure}
\end{minipage}
\begin{minipage}{0.02\linewidth}
\hfill
\end{minipage}
\begin{minipage}{0.48\linewidth}
\begin{subfigure}[t]{\linewidth}
\begin{small}
\vspace{1em}
\emphbox{0.99}{
\doubleemphbox{\textbf{Statements:}}

-- Suppose that [pay debts] is a part of [\doubleemphbox{forest}]

-- Suppose that [\doubleemphbox{forest}] \doubleemphbox{\orangeemph{does}} appear near [the planet]

\vspace{0.5em}

-- Suppose that [galaxy] is a part of [\doubleemphbox{coat}]

-- Suppose that [\doubleemphbox{coat}] does \doubleemphbox{\textcolor[HTML]{E06666}{\textbf{not}}} appear near [the planet]

\vspace{0.5em}

-- Suppose that [outer space] is a part of [\doubleemphbox{watch}]

-- Suppose that [\doubleemphbox{watch}] does \doubleemphbox{\textcolor[HTML]{E06666}{\textbf{not}}} appear near [the planet]

\vspace{0.5em}

-- Suppose that [orbit] is a part of [\doubleemphbox{story}]

-- Suppose that [\doubleemphbox{story}] does \doubleemphbox{\textcolor[HTML]{E06666}{\textbf{not}}} appear near [the planet]

\textbf{Question:}

Where is the first place someone leaving the planet ends up?

\textbf{Answer choices:}

A: \doubleemphbox{\orangeemph{pay debts}}\hspace{1em}B: galaxy\hspace{1em}C: outer space\hspace{1em}D: orbit
}
\end{small}
\caption{CSQA Augmented with \doubleemphbox{\AntiFactual{}} Reasoning}
\label{fig:csqa:af}
\end{subfigure}
\end{minipage}
\caption{The \accord{} framework \textbf{(b-g)} applied to a randomly-sampled CSQA instance \textbf{(a)}. Notice that \factual{} \textbf{(f)} and \antifactual{} \textbf{(g)} reasoning refers to whether the answer implied by the carefully-chosen \textcolor[HTML]{E06666}{\textbf{negation}} of the statements matches the original \factual{} answer \textbf{(a)}. The statements themselves are always \orangeemph{anti-factually} grounded.}
\label{fig:csqa}
\end{figure*}

\accord{} generates an \antifactual{} benchmark from an existing \factual{} commonsense dataset. In this work, we apply \accord{} to CommonsenseQA (CSQA) \cite{talmor2018commonsenseqa} to create the \accord$_\text{CSQA}$ benchmark suite. We chose CSQA due to its popularity.
However, \accord{} is
equally applicable to other commonsense reasoning datasets (see Appendix \ref{sec:appendix:other-datasets} for examples).

\accord's algorithm is complex. Figure \ref{fig:csqa} serves as a concrete running example throughout this Section. Significant additional detail and explanation is given in Appendices \ref{sec:appendix:accord}, \ref{sec:appendix:reductions} and \ref{sec:appendix:conceptnet}. In particular, Figure \ref{fig:accord} in Appendix \ref{sec:appendix:accord} illustrates an abstract view of the \accord{} framework using tree data structures. These two Figures are highly complementary and we encourage the reader to refer to both.

Consider a question-answer (QA) instance from CSQA with question $q$ and answer choices $\boldsymbol{a} = \{a_1, a_2, \ldots, a_Q\}$ (see Figure \ref{fig:csqa:original}). As discussed in \S\ref{sec:intro}, because CSQA is \wdef-aligned, LLMs can spuriously circumvent the reasoning required to arrive at the \factual{} answer \adef{} $\in \boldsymbol{a}$. To overcome this hurdle, we introduce an \antifactual{} context \caf, which consists of statements aligned to some \antifactual{} world, \waf{} $\ne$ \wdef{} (see Figure \ref{fig:csqa:factual}). Specifically, \caf{} is constructed to ensure some alternative answer choice \aaf{} $\ne$ \adef{} can also be implied via negation (see Figure \ref{fig:csqa:af}). The crux of \accord{} lies in the careful design of variants of \caf{} to systematically examine LLM performance (see Appendix \ref{sec:appendix:accord:motivation} for motivating design details).

\subsection{Reasoning Skills and Templates}
\label{sec:accord:reasoning_templates}

In order to rigorously quantify the differences between variants of \caf, \accord{} borrows from formal reasoning to partially formalize the vague and ill-defined underlying logic of commonsense reasoning \cite{huang2022towards, davis2023benchmarks} (see Figure \ref{fig:formal-vs-commonsense:formal-commonsense}). Thus, we construct \caf{} from formalized representations of so-called \textit{reasoning skills}.

\citet{talmor2018commonsenseqa} manually identified and defined reasoning skills needed to answer CSQA questions (see Table \ref{table:reasoning-skills-and-templates}), which we reuse for consistency. Each reasoning skill is a recurring pattern of commonsense reasoning. For example, the \texttt{spatial} reasoning skill is the pattern of understanding which objects commonsensically appear at which locations (e.g., ``an instrument appears near a music store''). CSQA derives from ConceptNet \cite{speer2017conceptnet}, an open graph of general commonsense knowledge. Thus, for each reasoning skill, we find a ConceptNet relation that is an appropriate and suitable match.\footnote{We omit the \texttt{social}, \texttt{activity}, and \texttt{definition} skills due to lack of a suitably matching ConceptNet relation.}

From each skill's definition, we devise a natural language \textit{reasoning template} that describes the corresponding reasoning skill. These templates (e.g., ``$\mathbf{X}$ appears near $\mathbf{Y}$'') serve as formalized representations of reasoning skills that link two concepts together via placeholder variables (e.g.,  $\mathbf{X} \leftarrow$ \texttt{instrument} and $\mathbf{Y} \leftarrow$ \texttt{music store}). We then compose these templates together to construct large \textit{reasoning trees} (see \S\ref{sec:accord:trees}), which form the basis of the formal structure underlying \caf.

\subsection{Pairing Templates}
\label{sec:accord:pairing_templates}

Not all reasoning skills are relevant to any particular CSQA instance. For example, the instance in Figure \ref{fig:csqa:original} is related to the \texttt{spatial} skill (e.g., ``What appears near the planet?''), but completely unrelated to, say, the \texttt{causal} skill. To ensure skill relevance, we craft \textit{pairing templates} for each CSQA instance (see Figure \ref{fig:csqa:template}). Pairing templates are special reasoning templates where one variable, the \textit{pairing variable}, $\mathbf{V}_p$, is grounded with a carefully-chosen question-specific term, the \textit{pairing term}, $p$. The other variable, $\mathbf{V}_x$, remains free. As discussed in \S\ref{sec:intro}, LLMs might spuriously circumvent \factual{} reasoning. To experimentally verify this, pairing templates are hand-crafted\footnote{Although pairing templates are hand-crafted, this human effort is, by design, highly reusable because a \textit{single} pairing template can pair a given CSQA instance to an \textit{arbitrary number} of reasoning trees of arbitrary complexity (see \S\ref{sec:accord:trees}).} such that they uniquely discriminate between the answer choices, $a_i \in \boldsymbol{a}$, of a given CSQA instance when setting $\mathbf{V}_x \leftarrow a_i$ and $\mathbf{V}_p \leftarrow p$.
In so doing, we allow the choice of $a_i$ to be either \factual{} or \antifactual{}, while also holding all other factors equal. 
Specifically, each pairing template has a \textit{positive variant}, $p^+$, which can imply any given answer choice, $a_i$, and a \textit{negative variant} (adding the ``not'' in Figure \ref{fig:csqa:template}), $p^-$ which can contradict $a_i$. See Appendix \ref{sec:appendix:accord:templates} for a detailed step-by-step example.

\subsection{Reasoning Trees}
\label{sec:accord:trees}

To construct a \textit{generic} reasoning tree, reasoning templates are composed (i.e., linked together) via their variables (see Appendix \ref{sec:appendix:accord:trees} for details). This creates a \textit{reasoning graph} with variables as nodes and templates as edges. \accord{} operates on this graph to rigorously enforce construct validity. In particular, \accord{} constrains the reasoning graph to contain no cycles---hence, the more precise term ``reasoning (poly)tree''. This acyclicity ensures reasoning trees do not entail reasoning paradoxes (e.g., ``$\mathbf{X}$ contains $\mathbf{Y}$; $\mathbf{Y}$ contains $\mathbf{Z}$; $\mathbf{Z}$ contains $\mathbf{X}$'').

In addition, two templates can link together only if they \textit{reduce} to a valid reasoning skill. Reducing two templates is algorithmically analogous to replacing two premises by their conclusion in logical deduction (see Figure \ref{fig:formal-vs-commonsense:formal}). Reducibility ensures that templates are composed commonsensically rather than arbitrarily, since not all combinations of two reasoning skills are reducible (see Appendix \ref{sec:appendix:reductions}). We pruned the set of all possible reasoning skill combinations to keep only those that are validly reducible. For example, whereas \texttt{part\_of} and \texttt{spatial} are reducible to \texttt{spatial} (see Figure \ref{fig:formal-vs-commonsense:formal}), \texttt{spatial} and \texttt{causal} are not---just because something is near a cause does not mean it is also near its effect. See Appendix \ref{sec:appendix:reductions} for details on all other combinations, including commonsense proofs of their validity.

As discussed in \S\ref{sec:accord:paths}, the difficulty level of \accord{} instances scales with tree size\footnote{Note that the trees shown throughout Figure \ref{fig:csqa} are kept small (tree size of 2) for legibility. See Figure \ref{fig:accord} for examples of more complex tree structures as the tree size increases.}. Since \accord{} automatically generates all possible valid generic trees up to a maximum size $T$ (chosen as a hyperparameter), \accord{} scales in difficulty with future LLM improvements using no additional human effort. This is atypical of commonsense reasoning benchmarks, which normally require human effort proportional to difficulty (see \S\ref{sec:related:scaling}).

Each CSQA instance is paired to only those trees matching one of its pairing templates (see Figure \ref{fig:csqa:generic_tree}), resulting in a subset of \textit{paired} reasoning trees (see Figure \ref{fig:csqa:paired_tree}) unique to each CSQA instance.

\subsection{Reasoning Paths}
\label{sec:accord:paths}

A \textit{reasoning path} is a sequence of linked templates in a paired reasoning tree that originate or terminate with a pairing template. The reasoning path connects a source variable with a (possibly distant) target variable. Due to the recursively reducible construction of reasoning trees (see Appendix \ref{sec:appendix:accord:paths}), $\mathbf{V}_p$ will necessarily be either the source or the target of the reasoning path. The variable at the other end is termed the \textit{answer variable}, $\mathbf{V}_a$ (see Figure \ref{fig:csqa:paired_tree}). Recall that grounding $\mathbf{V}_x \leftarrow a_i$ in a pairing template logically discriminates between the answer choices, $a_i \in \boldsymbol{a}$, of a CSQA instance (see \S\ref{sec:accord:pairing_templates}). By the nature of recursive reducibility, grounding $\mathbf{V}_a \leftarrow a_i$ necessarily does the same (see Appendix \ref{sec:appendix:accord:paths}). In a tree with $T$ templates, the \textit{reasoning complexity} is the number of templates, $n$, along a reasoning path.\footnote{This definition is chosen to align with the typical definition of \textit{reasoning hops} in the multi-hop reasoning literature (e.g., \citet{ho2020constructing}). We thus use the terms interchangeably.} Any templates outside a reasoning path are \textit{distractors}, $d$. Note that $T = n + d$ by construction. LLM performance in formal reasoning tasks tends to decrease as $n$ and $T$ increase \cite{xu2023large}, and is rarely robust to distractors \cite{kassner2019negated, wang2021adversarial, li2022large, misra2022comps, shi2023large}. As noted in \S\ref{sec:related}, \accord{} is unique among commonsense benchmarks in its ability to precisely quantify non-trivial $n$ and $d$ values, which empowers us to close the commonsense measurability gap by studying whether these trends apply here as well.

\subsection{Controlling Dataset Artifacts}

Since $\mathbf{V}_a$ can only be grounded with one $a_i \in \boldsymbol{a}$,
an LLM might ``guess'' that $a_i$ is the intended answer through simple lexical matching between \caf{} and $\boldsymbol{a}$, thereby spuriously circumventing the intended reasoning task via test-taking meta-reasoning \cite{mccoy2019right, chan2023spurious, li2023counterfactual}. To mitigate this potential bias, we duplicate the paired tree, once for \textit{each} $a_i$, which ensures each is present in \caf{} while holding all else equal (see Figure \ref{fig:csqa:duplication}). Several additional factors are employed to mitigate dataset artifacts. The order of the resulting statements, both within and between reasoning trees, is randomized to mitigate any potential systematic effects (such of recency bias). Duplicate statements are also removed, so as not to bias LLM reasoning towards them. These are not shown in Figure \ref{fig:csqa} only to aid with human legibility.

\subsection{\AntiFactual{} Grounding of Variables}
\label{sec:accord:grounding}

For each duplicated tree, the pairing and answer variables act as \textit{seed terms} from which we \orangeemph{anti-factually} ground all other variables using a backtracking beam tree search algorithm (see Appendix \ref{sec:appendix:conceptnet}). Since CSQA derives from ConceptNet, we employ it as our grounding KB. Essentially, since ConceptNet is a \factual{} KB, finding concept triples --- e.g., $\texttt{part\_of}(outer~space, watch)$ --- \textit{not} in ConceptNet enables us to argue that they are \antifactual. This relies on ConceptNet's recall: if a \factual{} triple is omitted from ConceptNet, we might mistake it as \antifactual. We mitigate against this using probabilistic hedging (see Appendix \ref{sec:appendix:conceptnet}).

\subsection{Selecting an Answer Using Negation}
\label{sec:accord:negation}

Finally, we negate the pairing template of all but one answer choice, $a_i$, to ensure that only $a_i$ is implied, while all other answer choices are contradicted. By carefully choosing which answer choice receives the positive variant, we control whether the statements in \caf{} together logically imply the \factual{} (see Figure \ref{fig:csqa:factual}) or some \antifactual{} (see Figure \ref{fig:csqa:af}) answer. Crucially, \textit{all other factors are held constant}, which enables direct comparison between \factual{}
and \antifactual{}
task variants (see Appendix \ref{sec:appendix:reductions} for additional details).

\section{Experiments}
\label{sec:experiments}

\begin{figure*}[th]
\centering
\begin{subfigure}{0.5425\linewidth}
\includegraphics[width=\linewidth]{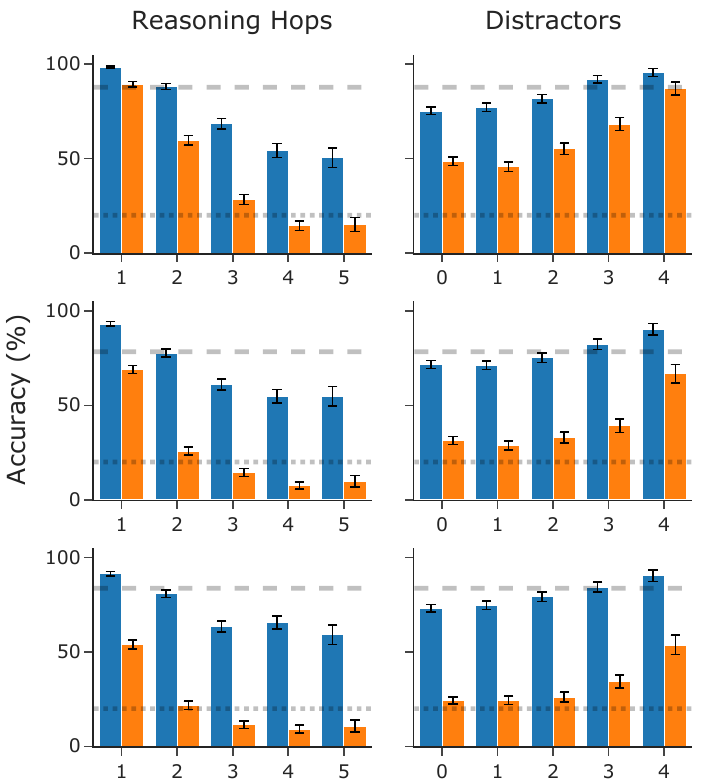}
\end{subfigure}
\begin{subfigure}{0.45\linewidth}
\includegraphics[width=\linewidth]{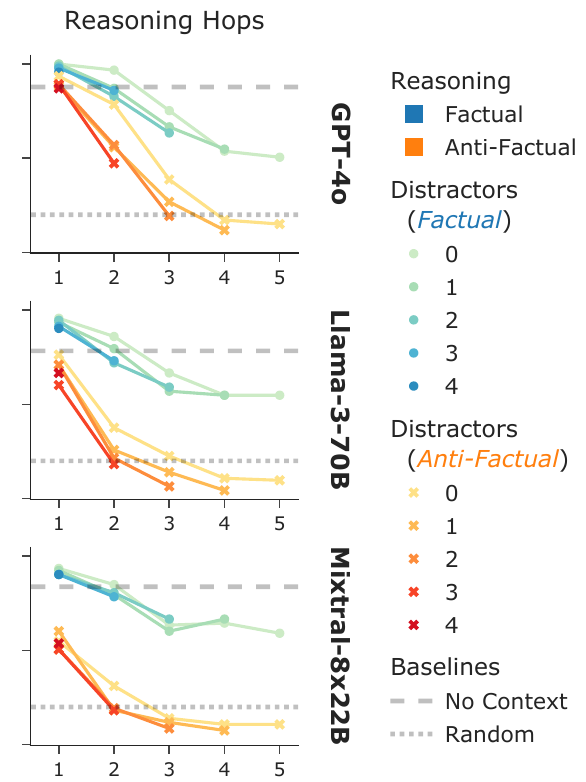}
\end{subfigure}
\caption{Performance of state-of-the-art LLMs on \accord$_\text{CSQA}$. \textbf{Left:} Both \factual{} and \antifactual{} performance degrade rapidly with increasing reasoning hops, which is expected. \textbf{Middle:} Both \factual{} and \antifactual{} performance increase with increasing distractors, which is unexpected. \textbf{Right:} Disentangling the interaction effect between reasoning hops and distractors to explain the unexpected result in \textbf{(Middle)}. Reasoning hops are dominant while distractors' effect is negligible, which explains the reversed trend in \textbf{(Middle)} when marginalizing over reasoning hops. \textbf{All:} \Factual{} significantly outperforms \antifactual, which indicates context unfaithfulness. As a consequence, \antifactual{} performance drops below random chance when reasoning hops exceed LLM reasoning capacity. Wald standard error bars are with respect to the 93 pairings, not reruns based on random seeds.}
\label{fig:results}
\end{figure*}

We apply \accord{} to CSQA to create \accord$_\text{CSQA}^N$, a benchmark suite where $N \in [0-5]$ represents the reasoning tree size. For each instance in the CSQA development set, we infer the ConceptNet relation that generated the instance by computing the majority vote among all ConceptNet assertions matching one of its answer choices to its source concept (see \citealt{talmor2018commonsenseqa} for details). The resulting distribution is highly skewed, with 607 instances of ConceptNet's \texttt{AtLocation} relation (the most popular) compared with 13 instances of \texttt{IsA} (the least popular). To balance \accord$_\text{CSQA}$, we randomly subsample each relation type, keeping only 13. Of the $6 \times 13 = 78$ instances, we reject two due to inciting violence, and another two where we cannot craft a valid pairing template. The baseline, \accord$_\text{CSQA}^0$, consists of the remaining 74 instances. We hand-craft 1$-$3 pairing templates for these 74 instances, resulting in 93 pairing templates. From these 93 pairings, we generate reasoning trees of sizes 1$-$5, producing a suite of benchmarks, \accord$_\text{CSQA}^{1-5}$, with 245,514 unique total trees and a guaranteed minimum of 143 unique trees per pairing. To reduce computational costs while ensuring wide coverage, we subsample the trees to keep precisely one tree per unique combination of reasoning hops and distractors per pairing, which produced a smaller benchmark with 2,864 unique trees. This step also balances the dataset, which is otherwise exponentially weighted towards the larger tree sizes (since there are exponentially more unique trees as the size increases). We present results from this smaller benchmark in Figure \ref{fig:results}, but release both versions (see Appendix \ref{sec:appendix:experiments} for additional details). 
Figure \ref{fig:results} shows the performance of GPT-4o (2024-05-13) \cite{gpt4o}, Llama-3-70B-Instruct \cite{llama3modelcard}, and Mixtral-8x22B-Instruct-v0.1 \cite{mixtralmodelcard} on \accord$_\text{CSQA}$ in a zero-shot setting. The ``No Context'' baseline represents \accord$_\text{CSQA}^0$.

\section{Results and Discussion}

Recall that, unlike prior commonsense benchmarks, \accord{} quantifies performance as a function of problem size, $T$, reasoning hops, $n$, and distractors, $d$, where $T=n+d$ (see \S\ref{sec:accord:paths}). Simultaneously, \accord{} disentangles (contextual) grounding and (parametric) reasoning effects through \textit{ceteris paribus} comparisons of \factual{} and \antifactual{} reasoning task variants (see \S\ref{sec:accord:negation}).

\paragraph{Are LLMs good reasoners or merely good parroters?}

For any given $n$ (Columns 1 and 3) or $d$ (Columns 2 and 3) in Figure \ref{fig:results}, \factual{} reasoning significantly outperforms \antifactual{} reasoning. Since we are carefully controlling for all other factors, this performance gap is directly indicative of context unfaithfulness: LLMs are inductively biased towards the \factual{} answer choice. In fact, \antifactual{} performance quickly degrades to \textit{below} random chance with only very moderate scaling of reasoning complexity. This suggests that as soon as their multi-hop reasoning capacity is exceeded, LLMs almost exclusively resort to spurious
shortcuts. Importantly, this performance gap is evidence of low construct validity in many \factual{} commonsense reasoning datasets: the intended measurement of LLM reasoning ability is overwhelmingly confounded by LLM parroting ability.

\paragraph{What is the effect of reasoning hops vs distractors?}

For $n>1$ in Column 1 of Figure \ref{fig:results}, both \factual{} and \antifactual{} performance degrade rapidly with increasing $n$. This trend has been understood in formal reasoning \cite{xu2023large}, but we demonstrate it here for the first time with commonsense. Interestingly, the trend for $d$ (Column 2) is reversed, which goes against expectations \cite{dalvi2021explaining, shi2023large}. Column 3 explains this artifact, which occurs because reasoning hops and distractors are complementary for a given problem size ($T=n+d$), and, as such, interact very strongly. In Column 3, when controlling for distractors, reasoning hops have a dominant effect on the trend. On the other hand, increasing distractors only slightly shifts the entire trend line downward. In other words, given a fixed $T$ (i.e., a fixed context budget), LLMs tend to prefer a smaller $n$ and larger $d$ than the reverse. That is, the ability of LLMs to filter out distractors significantly outclasses their multi-hop reasoning capacity. Marginalizing over reasoning hops (Column 2) obfuscates this nuance, resulting in a reversed trend. Being able to disentangle these effects is one of our key contributions.

\section{Related Work}
\label{sec:related}

\begin{table*}[t]
\centering
\begin{adjustbox}{max width=\linewidth}
\begin{tabular}{lccccccc}
\toprule
\textbf{Paper} & \textbf{CF} & \textbf{AF} & \textbf{Skills} & \textbf{Composable} & \textbf{Scalable} & \textbf{Hops} & \textbf{Distractors} \\
\midrule
2WikiMultiHopQA \cite{ho2020constructing} & \emptycirc & \emptycirc & \fullcirc & \halfcirc & \emptycirc & 2$-$? & ? \\
Fakepedia \cite{monea2023glitch} & \fullcirc & \fullcirc & \emptycirc & \halfcirc & \emptycirc & 1$-$2 & ? \\
DisentQA \cite{neeman2022disentqa} & \fullcirc & \fullcirc & \emptycirc & \emptycirc & \fullcirc & ? & ? \\
SCOTT \cite{wang2023scott} & \fullcirc & \fullcirc & \emptycirc & \emptycirc & \fullcirc & ? & ? \\
CConS \cite{kondo2023probing} & \fullcirc & \fullcirc & \halfcirc & \emptycirc & \emptycirc & 1 & 0 \\
CRASS \cite{frohberg2021crass} & \fullcirc & \halfcirc & \halfcirc & \emptycirc & \emptycirc & 1 & 0 \\
\midrule
\accord$_\text{CSQA}$ (Ours) & \fullcirc & \fullcirc & \fullcirc & \fullcirc & \fullcirc & \textbf{1$-$5} & \textbf{0$-$4} \\
\bottomrule
\end{tabular}
\end{adjustbox}
\caption{Comparison between \accord$_\text{CSQA}$ and related work. \textbf{CF}: Whether counterfactuals are present. \textbf{AF}: Whether those counterfactuals are specifically \antifactual. \textbf{Skills}: Whether reasoning is based on an explicit set of skills or on an uncontrolled process. \textbf{Composable}: Whether sub-components (e.g., templates) can be composed to generate more complex components. \textbf{Scalable}: Whether the reasoning complexity scales automatically---or, with at most $O(1)$ additional human effort. \textbf{Hops/Distractors}: Whether reasoning hops/distractors are measurable, and, if so, their range. \textbf{Legend}:\hspace{1.25em}\emptycirc[0.75ex] : no\hspace{1.5em}\halfcirc[0.75ex] : partially\hspace{1.5em}\fullcirc[0.75ex] : yes\hspace{1.5em}? : not measurable.}
\label{table:related-work}
\end{table*}

Table \ref{table:related-work} compares \accord{} to its most related commonsense reasoning benchmarks on key features.

\paragraph{Anti-Factual Counterfactuals.}

Counterfactual grounding is relevant for retrieval-augmented generation \cite{lewis2020retrieval, borgeaud2022improving, trivedi2022interleaving, gao2023retrieval, mialon2023augmented} and in-context knowledge editing \cite{zheng2023can}, as well as to context faithfulness more generally \cite{arodi2022kitmus, neeman2022disentqa, huang2023zero, ji2023rho, lanham2023measuring, sun2023towards, yu2023characterizing, zhou2023context}.
Counterfactuals in prior work can be hypothetical \cite{qin2019counterfactual, qin2020back, frohberg2021crass, wu2023reasoning} or \antifactual{} \cite{longpre2021entity, li2022large, neeman2022disentqa, kondo2023probing, monea2023glitch, tang2023large, wu2023reasoning, zhou2023context}, and can derive from diverse methodologies, including entity substitution in existing contexts \cite{longpre2021entity, li2022large, neeman2022disentqa, wu2023reasoning, zhou2023context}, generation using LLMs \cite{fu2023scene, monea2023glitch}, CoT demonstrations \cite{madaan2022text, wang2022towards, ye2022complementary, lanham2023measuring}, and other methods \cite{qin2019counterfactual, kaushik2020explaining, qin2020back, frohberg2021crass}. Entity substitution has garnered considerable attention for its simplicity in creating clear-cut \orangeemph{anti-factuals}. Unfortunately, it is  prone to lexical matching bias \cite{neeman2022disentqa, monea2023glitch}, which \accord{} avoids with tree duplication. In formal reasoning, LLM performance degrades when the surface forms of logic symbols are perturbed relative to their commonsense semantics \cite{dasgupta2022language, han2022folio, tang2023large, wu2023reasoning, yu2023ifqa}. \accord{} rigorously extends this analysis to commonsense.

\vspace{-0.5em}
\paragraph{Controlled Compositional Scaling.}
\label{sec:related:scaling}

Current LLMs struggle scaling to arbitrarily complex compositional and multi-hop reasoning tasks \cite{xu2023large, dziri2024faith}. Indeed, this may be a fundamental limitation of autoregressive architectures \cite{dziri2024faith}. Notwithstanding, existing commonsense tasks are limited to one- or two-hop reasoning \cite{ho2020constructing, frohberg2021crass, kondo2023probing, li2023counterfactual, davis2023benchmarks, monea2023glitch} or to an unknown number of hops and/or distractors \cite{chen2019understanding, kaushik2019learning, min2019compositional, ho2020constructing, longpre2021entity, neeman2022disentqa, davis2023benchmarks, monea2023glitch}. 
In commonsense reasoning, non-trivial \antifactual{} grounding is typically achieved only through manual effort, wherein each instance is handwritten. \accord{}, on the other hand, requires that only the pairing templates be handwritten. From these, arbitrarily many instances can be generated.
Such arbitrarily-scalable compositional reasoning is typical only of formal reasoning tasks \cite{dalvi2021explaining, tian2021diagnosing, han2022folio, tang2023large, xu2023large, dziri2024faith}. To the best of our knowledge, \accord{} is the first to introduce these features to commonsense reasoning. Furthermore, \accord{} carefully controls compositionality via a reasoning skill set, which is analogous to the controlled rule set of formal reasoning. Most prior commonsense benchmarks either are limited to a single skill---such as \texttt{spatial} \cite{kondo2023probing} or \texttt{causal} \cite{frohberg2021crass, li2023counterfactual, jin2024cladder}, or do not control skills \cite{neeman2022disentqa, davis2023benchmarks, monea2023glitch, wang2023scott}. To the best of our knowledge, only \citet{ho2020constructing} examine a rich set of highly-specific commonsense skills, such as \texttt{spouse}$(a, b)$ $\land$ mother$(b, c)$ $\Rightarrow$ \texttt{mother\_in\_law}$(a, c)$. This approach is complementary to \accord, which employs a rich set of general commonsense skills.

\section{Conclusion}

We presented \accord, a framework for generating \antifactual{} commonsense reasoning benchmarks. \accord{} introduces formal elements to commonsense reasoning, and thus takes a significant step towards closing the commonsense measurability gap with respect to formal reasoning. In particular, \accord{} disentangles commonsense grounding and reasoning abilities in LLMs, while controlling for both reasoning complexity, reasoning skills, and distractors. Experiments on our \accord$_\text{CSQA}$ benchmark suite, an application of the \accord{} framework to CSQA, demonstrate that the performance of state-of-the-art LLMs degrades to random chance with only moderate scaling, leaving substantial room for improvement. Moreover, we demonstrate a significant gap between \factual{} and \antifactual{} performance. This highlights the construct validity concerns of typical (\factual) benchmarks, which unfortunately allow LLMs to circumvent the intended reasoning task with parametric parroting of the \factual{} answer. \accord{} is uniquely designed to automatically scale its difficulty level in tandem with future LLM improvements by leveraging compositional scalability to generate future benchmarks of arbitrary reasoning complexity with minimal additional human effort.

\section*{Limitations}


\paragraph{Why have we not tried X or Y state-of-the-art prompting/fine-tuning technique?}

Our goal in this work is to introduce the \accord{} framework, present the rationale and development of the \accord$_\text{CSQA}$ benchmark suite, and illustrate baseline performance of LLMs in the simplest setting (zero-shot with a simple instruction prompt). We expect performance improvements with careful few-shot demonstrations and prompt engineering. We also expect that, through more detailed analysis than we presented here, \accord{} will yield significant additional insights both into LLM performance on commonsense reasoning tasks and into LLM context faithfulness on such tasks.  We welcome and encourage community adoption on these fronts!

\paragraph{Why do the instances in \accord{} resemble logic puzzles?}

Several prior works provide valuable detailed insight on template surface forms, such as the effects of verb and preposition choice \cite{kondo2023probing} or discourse connectives \cite{li2023counterfactual}. Since our reasoning skill set is significantly broader, we employed only one template per skill, while assuming that LLMs are increasingly able to abstract away template minutiae \cite{si2023measuring}.

As such, \accord's templates contain stilted language in practice, and the resulting benchmark instances may feel rather contrived to a human reader. This is a conscious trade-off. Specifically, in exchange for (a) automated scalability, (b) robust measurements of reasoning hops and distractors, and (c) assurances of context-faithful construct validity, our commonsense dataset superficially (but \textit{only} superficially) resembles a logic dataset.

Specifically, our goal is to measure, among other things, the reasoning complexity of commonsense problems. Because commonsense reasoning is vague and ill-defined, this cannot be done automatically (and therefore cannot be done scalably) in typical approaches to commonsense benchmark construction (unlike typical formal reasoning datasets). This is the commonsense measurability gap we address: Accurate and scalable measurements of commonsense reasoning complexity. The crux of our contribution is introducing formal elements borrowed from a more rigid type of reasoning (logic, in our case) to commonsense. The end product, therefore, \textit{resembles} a logic puzzle, but the \textit{crucial} difference is that a logic puzzle requires explicitly providing logical rules that must be followed. In our approach, the rules instead must be inferred commonsensically (see Figure \ref{fig:formal-vs-commonsense}). Hence, our work is a type of commonsense reasoning problem with formal logical elements added, rather than a type of logic problem with the formal rules removed.

\paragraph{The instances feel very unnatural and the lack of variation in the templates makes the instances feel stilted. Isn't that a problem?}

This is largely by design. Our \antifactual{} statements are designed to be fully solvable while also appearing unnatural enough at the surface level that LLMs cannot exploit their pre-trained world knowledge to circumvent the intended reasoning task. This forces the LLM to reason, rather than recite an answer from memory.

Unfortunately, improving the naturalness of \accord{} without sacrificing this benefit is non-trivial. Using LLMs as template editors to smooth out the stiltedness of the templates is challenging because LLMs may not faithful replicate the \antifactual{} grounding. For example, \citet{monea2023glitch} employ LLMs to generate counterfactuals from templates, but only about one-quarter of these are high-quality. Unfortunately, crowd workers are now widely employing LLMs to increase their productivity \cite{veselovsky2023artificial}, leading to the same problem.

Furthermore, variability in the templates may introduce confounds. We think that exploring ways to introduce the potential benefits of variability (e.g., question diversity) without also introducing potential costs (e.g., from confounds) is best left as future work, since it is not critical to the contributions we are making in this work and would instead distract from the main point (that LLMs recite more than they reason).

\paragraph{Why have we not included human assessments of quality? How is the benchmark validated? Would a human be expected to solve \accord{} tasks without issues?}

The quality of a typical formal reasoning dataset hinges on the quality of the underlying formal elements. For example, given a set of base propositions, axioms, and formal rules in a logical reasoning dataset, the quality of any given composed problem---regardless of complexity---is assured from the validity and soundness of these base elements. Analogously, the quality of \accord{} hinges on the validity and soundness of our formalizing elements (see Appendix \ref{sec:appendix:accord}) and especially our reduction matrix (see Appendix \ref{sec:appendix:reductions}). In traditional commonsense reasoning datasets, humans assess quality; but people are fallible, so most human-verified datasets---including CSQA---still contain errors \cite{davis2023benchmarks}. In our case, as is typical of logic datasets, the quality is instead assured algorithmically from the formal elements. We argue this results in a higher-quality dataset overall. Indeed, performance for \factual{} $n=1$ (Column 1 of Figure \ref{fig:results}) \textit{exceeds} the baseline for all three LLMs. We postulate this occurs because the noisiness of \textit{base} CSQA adds significant difficulty to the task \cite{kojima2022large, wei2022chain}, so having a single reasoning hop that reinforces the \factual{} answer choice
helps LLMs cut through that noise. In addition, humans, in general, should find \accord{} tasks as difficult as complex first-order logic tasks. As such, under the constrained time budget typically afforded on crowd worker platforms, humans would likely perform quite poorly on \accord, resulting in a poor assessment of its quality (see Appendix \ref{sec:appendix:human}).

\paragraph{Why are variables grounded with random concepts? Wouldn't it be better to use some similarity metric?}

Whereas ConceptNet enables us to control for relation types, it cannot control for the ``degree'' of \orangeemph{anti-factuality}. Consider, for example, that \textit{a fish is a type of mammal} intuitively feels ``less'' \antifactual{} than a \textit{fish is a type of rock}. We have attempted to formalize this intuition using both semantic distance between template concepts and likelihood functions of grounded templates \cite{monea2023glitch}. However, grounding a full reasoning path amounts to performing a random walk in template space. As a result, distant steps along a reasoning path can end up being more semantically related or more likely than all intermediate steps, which defeats the purpose of such a control mechanism. We leave this to future work.

\paragraph{Isn't \accord{} producing very noisy benchmarks?}

Most commonsense datasets and KBs are noisy, including CSQA and ConceptNet \cite{davis2023benchmarks}. For example, ConceptNet can be inconsistent with its relation types. Consider the following two assertions, both of which are in ConceptNet: \texttt{causal}(\texttt{playing lacrosse}, \texttt{fun}) and \texttt{used\_for}(\texttt{playing tennis}, \texttt{fun}). In both, playing a sport is related to having fun. In the first, lacrosse \textit{causes} fun, whereas in the latter, tennis is \textit{used for} fun. Both make commonsense. However, the inconsistency introduces noise when grounding \accord$_\text{CSQA}$ variables. While we ensure to always sample from the appropriate ConceptNet table based on the reasoning skill of the template for which we are grounding a variable, but cannot control for noise, mistakes, or inconsistencies within or across tables.

CSQA is also noisy. The majority of CSQA instances have inconsistent syntactic or semantic attributes (e.g., parts of speech, number, category, etc.) across their 5 answer choices, yet the associated crowd-sourced question is typically written such that only the correct answer matches all these attributes (e.g., see Figure \ref{fig:noisy_csqa} and Appendices \ref{sec:appendix:noise} \ref{sec:appendix:examples}). We attempted to minimize CSQA noise with carefully curated pairing templates that can differentiate the answer choices, even given question and answer choice noise. However, we were unable to craft valid pairing templates for two of the 78 base CSQA instances (see Appendix \ref{sec:appendix:noise}).

Importantly, CSQA noise hardly limits the quality of \accord$_\text{CSQA}$, because our benchmark includes both \factual{} and \antifactual{} conclusions for each question. The critical measure of overall performance is the \textit{difference} between \factual{} and \antifactual{} LLM performance on each question, rather than the absolute performance level on either task. Since we control for the CSQA question, both the \factual{} and \antifactual{} variants have the same noisiness. As such, we are effectively controlling for much of CSQA's noisiness in our experiments.

In fact, performance for \factual{} $n=1$ (Column 1) exceeds this baseline for all three LLMs. We postulate this occurs because the noisiness of \textit{base} CSQA adds significant difficulty to the task \cite{kojima2022large, wei2022chain}, yet having a \textit{single} reasoning hop that reinforces the \factual{} answer choice in the most straightforward manner possible helps LLMs cut through that noise.

\section*{Acknowledgements}
Resources used in preparing this research were provided, in part, by the Province of Ontario, the Government of Canada through CIFAR, and companies sponsoring the Vector Institute \url{www.vectorinstitute.ai/\#partners}. Frank Rudzicz is supported by a Canada CIFAR Chair in AI. We would like to thank Ian Berlot-Attwell and Kalyna Kryworuchko for their valued insights on this research.

\bibliography{custom}

\appendix

\section{Applying \accord{} to Other Datasets}
\label{sec:appendix:other-datasets}

We designed \accord{} to generalize beyond CSQA. Indeed, the preprocessing steps (top row of Figure \ref{fig:accord}) are used to \textit{mold} existing datasets or benchmarks to \accord’s agnostic framework. They are not part of \accord{} itself. We also limited \accord$_\text{CSQA}$ to the 6 reasoning skills identified by \citet{talmor2018commonsenseqa} for which we could find equivalent ConceptNet relations, which limits our reduction matrix (see Appendix \ref{sec:appendix:reductions}) to those 6 skills. However, \accord{} can generalize beyond these as well.

Although we do not implement any additional benchmarks beyond CSQA in this work, here we outline illustrative examples applying \accord{} to both ARC \cite{clark2018think} and OpenBookQA \cite{mihaylov2018can}.

\paragraph{ARC:}

Convert Table 4 into new reasoning skills. Some of these overlap with CSQA's 6 reasoning skills, but many are novel. Table 5 lists question types. Use the ``Multihop Reasoning'' category as a basis from which to create \antifactual{} instances. Avoid the ``Hypothetical / Counterfactual'' category entirely. Instead of ConceptNet as the grounding KB, use the ARC Corpus (which contains sentences) together with the related WorldTree KB \cite{jansen2018worldtree} (which converts ARC-related sentences into structured graphs).

\paragraph{OpenBookQA:}

Figure 1 illustrates the dataset construction: Science facts pulled from WorldTree \cite{jansen2018worldtree} as well implicit common knowledge. Make these common knowledge facts explicit by turning them into reasoning templates, then make them \antifactual{} (e.g., substituting `metal' with another entity lacking the central property). This could be done using a property-aware KB, like WikiData \cite{vrandevcic2014wikidata} or WorldTree. Table 2 is a starting point for reasoning skills and Table 3 for reductions. In fact, the ``Reasoning Challenge'' column of Table 3 maps cleanly to our reduction proofs (see Appendix \ref{sec:appendix:reductions}).

\section{Additional Details of the \accord{} Algorithm}
\label{sec:appendix:accord}

\begin{figure*}[t]
\centering
\includegraphics[width=0.95\linewidth]{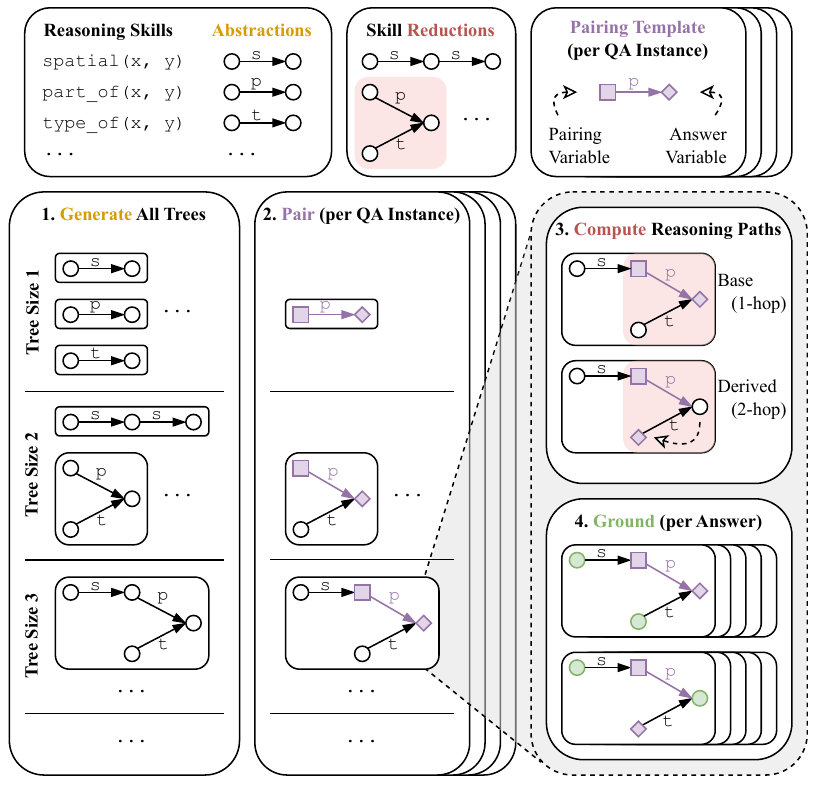}
\caption{The \accord{} framework applied to CSQA. \textbf{Top row:} Manual preprocessing of CSQA. \textbf{Bottom:} Fully automated steps based on this preprocessing. \textbf{(1)} Generate all possible reasoning trees. \textbf{(2)} Pair each CSQA instance to all matching trees. \textbf{(3)} Find all $n$-hop reasoning paths to vary the number of reasoning hops and distractors. \textbf{(4)} For each path, duplicate the tree for each answer choice, then \orangeemph{anti-factually} ground variables. \textbf{Legend:} \texttt{s}, \texttt{p}, \texttt{t} in the abstractions are shorthands for \texttt{spatial}, \texttt{part\_of}, and \texttt{type\_of}, respectively.}
\label{fig:accord}
\end{figure*}

Figure \ref{fig:accord} presents an abstracted view of the \accord{} framework. In this Section, we will provide additional \accord{} details using both Figures \ref{fig:csqa} and \ref{fig:accord} as reference.

\subsection{Motivating Design Argument}
\label{sec:appendix:accord:motivation}

\begin{figure*}[t]
\centering
\begin{minipage}{0.48\linewidth}
\begin{subfigure}[t]{\linewidth}
\begin{small}
\emphbox{0.99}{
\textbf{Pairing Template:}

-- Suppose that [\doubleemphbox{outer space}] \doubleemphbox{\blueemph{does}} appear near [the planet]

\textbf{Question:}

Where is the first place someone leaving the planet ends up?

\textbf{Answer choices:}

A: pay debts\hspace{1em}B: galaxy\hspace{1em}C: \doubleemphbox{\blueemph{outer space}}\hspace{1em}D: orbit
}
\end{small}
\caption{Pairing Template Implies \Factual{} Answer}
\label{fig:pairing_templates:factual_implies}
\end{subfigure}
\end{minipage}
\begin{minipage}{0.02\linewidth}
\hfill
\end{minipage}
\begin{minipage}{0.48\linewidth}
\begin{subfigure}[t]{\linewidth}
\begin{small}
\emphbox{0.99}{
\textbf{Pairing Template:}

-- Suppose that [\doubleemphbox{outer space}] does \doubleemphbox{\textcolor[HTML]{E06666}{\textbf{not}}} appear near [the planet]

\textbf{Question:}

Where is the first place someone leaving the planet ends up?

\textbf{Answer choices:}

A: pay debts\hspace{1em}B: galaxy\hspace{1em}C: outer space\hspace{1em}D: orbit
}
\end{small}
\caption{Pairing Template Contradicts \Factual{} Answer}
\label{fig:pairing_templates:factual_contradict}
\end{subfigure}
\end{minipage}
\begin{minipage}{0.48\linewidth}
\begin{subfigure}[t]{\linewidth}
\vspace{1em}
\begin{small}
\emphbox{0.99}{
\textbf{Pairing Template:}

-- Suppose that [\doubleemphbox{pay debts}] \doubleemphbox{\orangeemph{does}} appear near [the planet]

\textbf{Question:}

Where is the first place someone leaving the planet ends up?

\textbf{Answer choices:}

A: \doubleemphbox{\orangeemph{pay debts}}\hspace{1em}B: galaxy\hspace{1em}C: \blueemph{outer space}\hspace{1em}D: orbit
}
\end{small}
\caption{Pairing Template Implies \AntiFactual{} Answer}
\label{fig:pairing_templates:af_implies}
\end{subfigure}
\end{minipage}
\begin{minipage}{0.02\linewidth}
\hfill
\end{minipage}
\begin{minipage}{0.48\linewidth}
\begin{subfigure}[t]{\linewidth}
\vspace{1em}
\begin{small}
\emphbox{0.99}{
\textbf{Pairing Template:}

-- Suppose that [\doubleemphbox{pay debts}] does \doubleemphbox{\textcolor[HTML]{E06666}{\textbf{not}}} appear near [the planet]

\textbf{Question:}

Where is the first place someone leaving the planet ends up?

\textbf{Answer choices:}

A: pay debts\hspace{1em}B: galaxy\hspace{1em}C: \blueemph{outer space}\hspace{1em}D: orbit
}
\end{small}
\caption{Pairing Template Contradicts \AntiFactual{} Answer}
\label{fig:pairing_templates:af_contradict}
\end{subfigure}
\end{minipage}
\caption{Manipulating a pairing template to imply or contradict various answer choices.}
\label{fig:pairing_templates}
\end{figure*}

Our goal is this work is to close the commonsense measurability gap with respect to formal reasoning. Our design objectives, therefore, are: (1) to be able to explicitly, rigorously, and automatically quantify the reasoning complexity of a commonsense task; and (2) to mitigate construct validity concerns stemming from context unfaithfulness and spurious shortcuts in LLMs (see \S\ref{sec:intro}).

Consider the example CSQA instance from Figure \ref{fig:csqa:original}. How do we quantify this instance's reasoning complexity? Our proposed solution is to add a context containing reasoning statements that, together with an implicit commonsense understanding of the underlying rules, imply the original answer. By carefully controlling the automated construction of these statements, we can explicitly, rigorously, and automatically quantify reasoning complexity. The task for an LLM is to reason through the statements using its understanding of these rules to derive the answer.

Notice, however, that the LLM can just as easily ignore this context and instead answer the question directly from parametric memory. How do we detect such context unfaithfulness to maintain construct validity? Our proposed solution is to design the context in such a way that it can imply an alternative answer choice (which we term the \antifactual{} reasoning task variant) in addition to the original answer (which we term the \factual{} reasoning task variant) given the same CSQA instance. In the \factual{} reasoning task, the LLM can arrive at the desired answer either by reasoning through the context as intended or by spuriously parroting the answer while ignoring the context. In the \antifactual{} reasoning task, the LLM can arrive at the desired answer only by reasoning through the context.\footnote{The LLM can also arrive at the desired \antifactual{} answer by attempting to spuriously guess the \factual{} answer, but guessing incorrectly and thereby accidentally landing on the desired \antifactual{} answer. Empirically, the fact that LLMs perform \textit{worse} than random chance (see Figure \ref{fig:results}) demonstrates that this occurs only very rarely, if at all.} As such, the gap between the \factual{} and \antifactual{} performance is directly indicative of context unfaithfulness in LLMs and highlights the construct validity concern in the \factual{} reasoning task, where the desired measurement (reasoning ability) is confounded (by parroting ability).

Since a purely \factual{} context cannot imply a \blueemph{factually} incorrect answer, to create an \antifactual{} reasoning task, the context must itself be \orangeemph{anti-factually} grounded. On the other hand, the context for the \factual{} reasoning task can be either \blueemph{factually} or \orangeemph{anti-factually} grounded. This works because an \antifactual{} context can imply any answer choice (including the \factual{} answer) with appropriate negation (compare Figures \ref{fig:csqa:factual} and \ref{fig:csqa:af}).

However, there are three arguments in favour of using an \antifactual{} context even with the \factual{} reasoning task. First, a purely \factual{} context can be spuriously circumvented, thereby reducing the intended reasoning complexity of the task. For example, from the context ``$\mathbf{A}$ appears near $\mathbf{B}$; $\mathbf{B}$ appears near $\mathbf{C}$'', we can derive ``$\mathbf{A}$ appears near $\mathbf{C}$''. If the derivation is \factual, an LLM might be able to spuriously retrieve it without reasoning through the context. The derivation could then be used to directly answer the question, turning a 2-hop reasoning task into a 1-hop task. Second, a stronger direct comparison can be made between the \factual{} and \antifactual{} task variants if all else is being held equal except the answer choice selection (compare Figures \ref{fig:csqa:factual} and \ref{fig:csqa:af}). Third, empirically, it is challenging to \blueemph{factually} ground $n$-hop contexts with $n>2$ due to the sparsity of grounding KBs. ConceptNet, for example, simply does not contain enough many-hop chains to ground even a \textit{single} 3-hop instance from our benchmark.

In sum, to be able to explicitly, rigorously, and automatically quantify the reasoning complexity of a commonsense task, and to mitigate construct validity concerns stemming from context unfaithfulness and spurious shortcuts in LLMs, we design an \orangeemph{anti-factually} grounded context, \caf, with which to augment CSQA instances such that any answer choice can be implied while holding all else equal. The performance gap between a context that implies the \factual{} vs some \antifactual{} answer is directly indicative of context unfaithfulness in LLMs. The mere presence of \caf{} renders the reasoning complexity explicit. The details of the careful automated construction and validation of \caf{} to ensure rigour is discussed below.

\subsection{Pairing Templates}
\label{sec:appendix:accord:templates}

In light of the discussion in Appendix \ref{sec:appendix:accord:motivation}, our goal is to develop an algorithm that automates: (a) the construction of an \antifactual{} context, \caf; and (b) the rigorous validation of \caf's soundness.

Consider the example CSQA instance from Figure \ref{fig:csqa:original}. How do we devise a simple context that can discriminate between its answer choices? Our proposed solution is the \textit{pairing template}. Each pairing template is a statement with two placeholder variables. The goal is to craft the pairing template such that, if one of these variables is grounded with a particular answer choice $a_i$, and the other variable is grounded with a fixed shared term between all answer choices (the \textit{pairing term}), then the statement as a whole can directly imply (using the positive variant of the pairing template) or contradict (using the negative variant) $a_i$. The template must also satisfy the form of a reasoning template. In particular, it must derive from one of the allowable reasoning skills (see Table \ref{table:reasoning-skills-and-templates}).

Figure \ref{fig:pairing_templates} provides a detailed example of implication and contradiction of either the \factual{} or some \antifactual{} answer choice for the running example from Figure \ref{fig:csqa}. Notice that `implying' an \antifactual{} answer requires assuming some \antifactual{} world model, \waf. Hence, we craft the pairing template to begin with ``Suppose that''. Conversely, `contradicting' the \factual{} answer requires assuming the \blueemph{default} world model is wrong (again with ``Suppose that''). Only by combining these assumption using tree duplication and negation (discussed below) can the \antifactual{} rightly be considered implied.

\subsection{Reasoning Trees}
\label{sec:appendix:accord:trees}

Each \textit{reasoning template} represents one reasoning hop. Composing reasoning templates together into \textit{reasoning (poly)trees} therefore creates multi-hop reasoning objects. In the same way that a formal algorithm can operate rigorously on formal reasoning objects, such as an algebraic equation or a first-order logic problem, we devise an algorithm (\accord) to operate rigorously on these trees. \accord{} not only generates all possible trees of a certain reasoning complexity, but also ensures reasoning soundness and validity of the resulting object. In essence, we are applying formal reasoning rules to commonsense reasoning problems to automate both problem generation and verification.

Figure \ref{fig:accord} illustrates the main steps of \accord. We assume as input a set of reasoning skills and a corresponding reduction matrix. Skill reductions are discussed in detail in Appendix \ref{sec:appendix:reductions}. For now, we can treat this reduction matrix as a lookup table: Given two reasoning skills, the matrix tells us whether these skills can be simplified or not, and what the skill type of the resulting simplification is (if any). By including only reductions that are closed under the set of skills, we are able to recursively generate trees of arbitrary size.

Trees of size 1 are degenerate: There is precisely one such tree for each reasoning skill (shown as a node-and-edge tree abstraction in Figure \ref{fig:accord}). Trees of size 2 comprise precisely the set of reductions from the reduction matrix (since the reduction matrix, by definition, contains exactly those pairs of reasoning skills that are valid).

As an example of a tree of size 3, consider the tree under the heading ``Tree Size 3'' in Figure \ref{fig:accord}. To construct this tree, \accord{} arbitrarily starts with one reasoning skill, say \texttt{type\_of}. Then, using the reduction matrix, it finds that $\texttt{part\_of}(x,y) \land \texttt{type\_of}(z,y)$ form a valid reduction to $\texttt{part\_of}(x,z)$, thereby growing the tree. Next, \accord{} arbitrarily picks a branch in the tree, say \texttt{part\_of}, and attempts to grow the tree. It finds that $\texttt{part\_of}(x,y) \land \texttt{spatial}(y,z)$ reduce to $\texttt{spatial}(x,z)$. \accord{} then performs some additional sanity checks to ensure validity, such as strictly enforcing acyclicity and using variable substitution to ensure unique variable names, then spits out this completed 3-hop tree before recursively backtracking to generate all other valid trees.

\subsection{Reasoning Paths}
\label{sec:appendix:accord:paths}

\begin{figure}[t]
\centering
\begin{subfigure}[t]{\linewidth}
\begin{small}
\emphbox{0.99}{
\textbf{Statements:}

-- Suppose that [\doubleemphbox{outer space}] appears near [the planet]

\textbf{Question:}

Where is the first place someone leaving the planet ends up?

\textbf{Answer choices:}

A: pay debts\hspace{1em}B: galaxy\hspace{1em}C: \blueemph{outer space}\hspace{1em}D: orbit
}
\end{small}
\caption{Tree Size 1; Reasoning Hops 1; Distractors 0}
\label{fig:paths:one}
\end{subfigure}
\begin{subfigure}[t]{\linewidth}
\begin{small}
\emphbox{0.99}{
\textbf{Statements:}

-- Suppose that [\doubleemphbox{outer space}] appears near [the planet]

-- Suppose that [\doubleemphbox{the planet}] appears near [\doubleemphbox{Jupiter}]

\textbf{Question:}

Where is the first place someone leaving the planet ends up?

\textbf{Answer choices:}

A: pay debts\hspace{1em}B: galaxy\hspace{1em}C: \blueemph{outer space}\hspace{1em}D: orbit
}
\end{small}
\caption{Tree Size 2; Reasoning Hops 1; Distractors 1}
\label{fig:paths:dist}
\end{subfigure}
\begin{subfigure}[t]{\linewidth}
\vspace{1em}
\begin{small}
\emphbox{0.99}{
\textbf{Statements:}

-- Suppose that [\doubleemphbox{outer space}] is a part of [\doubleemphbox{watch}]

-- Suppose that [\doubleemphbox{watch}] appears near [the planet]

\textbf{Question:}

Where is the first place someone leaving the planet ends up?

\textbf{Answer choices:}

A: pay debts\hspace{1em}B: galaxy\hspace{1em}C: \blueemph{outer space}\hspace{1em}D: orbit
}
\end{small}
\caption{Tree Size 2; Reasoning Hops 2; Distractors 0}
\label{fig:paths:two}
\end{subfigure}
\begin{subfigure}[t]{\linewidth}
\vspace{1em}
\begin{small}
\emphbox{0.99}{
\textbf{Statements:}

-- Suppose that [\doubleemphbox{outer space}] is a part of [\doubleemphbox{watch}]

-- Suppose that [\doubleemphbox{watch}] appears near [the planet]

-- Suppose that [\doubleemphbox{the planet}] appears near [\doubleemphbox{Jupiter}]

\textbf{Question:}

Where is the first place someone leaving the planet ends up?

\textbf{Answer choices:}

A: pay debts\hspace{1em}B: galaxy\hspace{1em}C: \blueemph{outer space}\hspace{1em}D: orbit
}
\end{small}
\caption{Tree Size 3; Reasoning Hop 2; Distractors 1}
\label{fig:paths:two_dist}
\end{subfigure}
\caption{The logic of reductions paths ensures that all these examples validly imply ``outer space'' regardless of tree size, reasoning hops, or number of distractors.}
\label{fig:paths}
\end{figure}

Not all generic reasoning trees are relevant to the commonsense underlying any particular CSQA instance. The pairing template serves as an anchor point between generic trees and a CSQA instance. Only those trees containing a reasoning skill matching the skill type of the pairing template can be paired to that CSQA instance. If the tree contains multiple instances of the same reasoning skill, each instance can pair separately with the CSQA instance. In essence, the specifics of the pairing template ``replace'' the generic reasoning skill in the tree, resulting in a \textit{paired reasoning tree}.

From this pairing template anchor point in the tree, \accord{} then computes all valid \textit{reasoning paths}. Specifically, \accord{} looks for skill branches in the tree adjacent to the designated pairing template that, when combined with the pairing template reduce to the \textbf{same} skill type as the pairing template. Moreover, the \textit{pairing term} variable must \textbf{survive} the reduction (it cannot be an intermediate variable that gets lost through the reduction).

As an example, consider the reduction of $\texttt{part\_of}(x,y) \land \texttt{spatial}(y,z)$ to $\texttt{spatial}(x,z)$. Because the skill type of the reduction is \texttt{spatial}, the path [\texttt{part\_of} $\rightarrow$ \texttt{spatial}] is only valid if \texttt{spatial} is the pairing template (if \texttt{part\_of} happened to be the pairing template, this tree branch is rejected as a possible reasoning path because the pairing template's skill type does not match that of the reduction). Moreover, the reasoning path is only valid if either $x$ or $z$ is the pairing term variable, since these survive the reduction ($y$ does not, so this reasoning path is rejected if $y$ happened to be the pairing term).

Together, these two conditions ensure that the pairing template \textbf{dominates} the reasoning path. Dominance is important, because a pairing template can \textit{only} be in a dominant position in a reasoning path. Otherwise, its answer-discrimination ability becomes subsumed. Essentially, finding a reasoning path is logically equivalent to applying the reduction, then removing the original skill branches from the reasoning tree and replacing them with the reduction. This simplifies the reasoning complexity of the problem by one hop while ensuring that the pairing template continues to appear in the reduced tree. Since the pairing template validly matches the reduced tree, we have not altered the essence of the reasoning logic, only its complexity. Recursively applying such reductions results in a maximally reduced tree containing exactly one reasoning hop (the dominant pairing template) and, optionally, some irrelevant statements (distractors). We can also stop the recursive reduction process short in order to create intermediate-length reasoning paths.

Because the pairing template dominates throughout, and because the pairing template is designed to uniquely discriminate between the different answer choices $a_i$ (see Figure \ref{fig:pairing_templates}), we know that the full reasoning path also necessarily discriminates between the different answer choices when grounding the variable at the opposite end of the reasoning path from the pairing template with $a_i$.

Consider our running example. With a 1-hop tree (which is equivalent to just the pairing template), we know that the tree as a whole can discriminate between the answer choices (see Figure \ref{fig:pairing_templates}). In Figure \ref{fig:paths:one}, we reproduce this 1-hop tree example. In Figure \ref{fig:paths:dist}, we instead introduce a reasoning tree of size 2. This tree continues to have only one reasoning hop, however, since the other statement is a distractor. This is because both the \textit{pairing term} and the \textit{answer choice} ground variables in the \textit{same} statement. Notice that we can completely ignore the distracting statement, since it has no bearing on the reasoning logic on the problem. In Figure \ref{fig:paths:two}, we introduce a different reasoning tree of size 2. This reasoning tree has two reasoning hops, because there are two statements between the \textit{pairing term} and \textit{answer choice} variables. Notice that, because we know that $\texttt{part\_of}(x,y) \land \texttt{spatial}(y,z)$ reduce to $\texttt{spatial}(x,z)$, and because we know that the pairing template (\texttt{spatial}) dominates the other statement (\texttt{part\_of}), we can be confident in knowing that the 2-hop statements imply the answer choice.\footnote{Assuming an \antifactual{} world model, \waf, if it is true that outer space is part of a watch and that a watch appears near the planet, then we can conclude that outer space is the correct answer to the question.} Finally, in Figure \ref{fig:paths:two_dist}, we combine the previous two cases: A tree of size 3 with two reasoning hops and one distractor. This case combines the logic of the others: The 2-hop reasoning is valid and the distractor can be ignored. 

In Figure \ref{fig:accord}, we show more abstractly how a reasoning path can be computed through a size 3 reasoning tree. Notice that, in this example, \texttt{part\_of} is the pairing template. Because \texttt{spatial} dominates \texttt{part\_of} in the reduction matrix, the reasoning path cannot more ``backwards'' to include \texttt{spatial}. On the other hand, \texttt{part\_of} does dominate \texttt{type\_of} (the reduced skill type remains \texttt{part\_of} and the pairing term variable survives the reduction). As such, the reasoning path can be extended ``downwards'' to include \texttt{type\_of}. Notice that the answer choice variable ``moves'' to the end of the reasoning path, whereas the pairing term variable remains fixed to the pairing template.

\section{Reduction Matrix and Informal Proofs}
\label{sec:appendix:reductions}

\begin{figure}[t]
\centering
\begin{subfigure}[t]{\linewidth}
\begin{adjustbox}{max width=\linewidth}
\begin{tabular}{|p{0.5em}|p{0.5em}|@{\hspace{-0.5\arrayrulewidth}}c@{\hspace{-0.5\arrayrulewidth}}|p{0.5em}|p{0.5em}|@{\hspace{-0.5\arrayrulewidth}}c@{\hspace{-0.5\arrayrulewidth}}|p{0.5em}|p{0.5em}|@{\hspace{-0.5\arrayrulewidth}}c@{\hspace{-0.5\arrayrulewidth}}|p{0.5em}|p{0.5em}|@{\hspace{-0.5\arrayrulewidth}}c@{\hspace{-0.5\arrayrulewidth}}|p{0.5em}|p{0.5em}|@{\hspace{-0.5\arrayrulewidth}}c@{\hspace{-0.5\arrayrulewidth}}|p{0.5em}|p{0.5em}|l}
\rot{\texttt{spatial}} & \rot{\texttt{causal}} & \rot{\texttt{part\_of}} & \rot{\texttt{type\_of}} & \rot{\texttt{used\_for}} & \rot{\texttt{requires}} \\
\hhline{*{6}{|--|~}}
& \cellcolor[HTML]{93C47D} \texttt{s} & \makebox[\doublerulesep]{} & & & \makebox[\doublerulesep]{} & & \cellcolor[HTML]{93C47D} \texttt{s} & \makebox[\doublerulesep]{} & \cellcolor[HTML]{93C47D} \texttt{s} & & \makebox[\doublerulesep]{} & & & \makebox[\doublerulesep]{} & & & \multirow{2}{*}{\texttt{s}} \\
\hhline{-|-|~*{5}{|--|~}}
\multicolumn{1}{c|}{} & & \makebox[\doublerulesep]{} & & & \makebox[\doublerulesep]{} & \cellcolor[HTML]{93C47D} \texttt{s} & & \makebox[\doublerulesep]{} & \cellcolor[HTML]{93C47D} \texttt{s} & & \makebox[\doublerulesep]{} & & & \makebox[\doublerulesep]{} & & & \\
\cline{2-2}\hhline{*{1}{~~~}*{5}{:==:~}}
\multicolumn{3}{c|}{} & & \cellcolor[HTML]{93C47D} \texttt{c} & \makebox[\doublerulesep]{} & & & \makebox[\doublerulesep]{} & \cellcolor[HTML]{93C47D} \texttt{c} & & \makebox[\doublerulesep]{} & & \cellcolor[HTML]{93C47D} \texttt{u} & \makebox[\doublerulesep]{} & & & \multirow{2}{*}{\texttt{c}} \\
\hhline{*{1}{~~~}-|-|~*{4}{|--|~}}
\multicolumn{4}{c|}{} & & \makebox[\doublerulesep]{} & & & \makebox[\doublerulesep]{} & \cellcolor[HTML]{93C47D} \texttt{c} & & \makebox[\doublerulesep]{} & & & \makebox[\doublerulesep]{} & & & \\
\cline{5-5}\hhline{*{2}{~~~}*{4}{:==:~}}
\multicolumn{6}{c|}{} & & \cellcolor[HTML]{93C47D} \texttt{p} & \makebox[\doublerulesep]{} & \cellcolor[HTML]{93C47D} \texttt{p} & & \makebox[\doublerulesep]{} & & & \makebox[\doublerulesep]{} & & & \multirow{2}{*}{\texttt{p}} \\
\hhline{*{2}{~~~}-|-|~*{3}{|--|~}}
\multicolumn{7}{c|}{} & & \makebox[\doublerulesep]{} & & & \makebox[\doublerulesep]{} & & & \makebox[\doublerulesep]{} & & & \\
\cline{8-8}\hhline{*{3}{~~~}*{3}{:==:~}}
\multicolumn{9}{c|}{} & & \cellcolor[HTML]{93C47D} \texttt{t} & \makebox[\doublerulesep]{} & & \cellcolor[HTML]{93C47D} \texttt{u} & \makebox[\doublerulesep]{} & & \cellcolor[HTML]{93C47D} \texttt{r} & \multirow{2}{*}{\texttt{t}} \\
\hhline{*{3}{~~~}-|-|~*{2}{|--|~}}
\multicolumn{10}{c|}{} & & \makebox[\doublerulesep]{} & & & \makebox[\doublerulesep]{} & & & \\
\cline{11-11}\hhline{*{4}{~~~}*{2}{:==:~}}
\multicolumn{12}{c|}{} & & \cellcolor[HTML]{93C47D} \texttt{u} & \makebox[\doublerulesep]{} & \cellcolor[HTML]{93C47D} \texttt{u} & & \multirow{2}{*}{\texttt{u}} \\
\hhline{*{4}{~~~}-|-|~*{1}{|--|~}}
\multicolumn{13}{c|}{} & & \makebox[\doublerulesep]{} & & & \\
\cline{14-14}\hhline{--~*{4}{~~~}*{1}{:==:~}}
{\tiny $>$} & {\tiny $\rightarrow$} & \multicolumn{13}{l|}{\textbf{Permutation}} & & \cellcolor[HTML]{93C47D} \texttt{r} & \multirow{2}{*}{\texttt{r}} \\
\hhline{|--|~*{4}{~~~}-|-|~}
{\tiny $\leftarrow$} & {\tiny $<$} & \multicolumn{14}{l|}{\textbf{Legend}} & & \\
\cline{1-2}\cline{17-17}
\end{tabular}
\end{adjustbox}
\end{subfigure}
\begin{subfigure}[b]{\linewidth}
\vspace{0.5em}
\begin{adjustbox}{max width=\linewidth}
\textbf{Case $>$:} $\texttt{row}(x,y) \land \texttt{col}(z,y) \Rightarrow \texttt{reduction}(x,z)$
\end{adjustbox}
\begin{adjustbox}{max width=\linewidth}
\textbf{Case $\rightarrow$:} $\texttt{row}(x,y) \land \texttt{col}(y,z) \Rightarrow \texttt{reduction}(x,z)$
\end{adjustbox}
\begin{adjustbox}{max width=\linewidth}
\textbf{Case $\leftarrow$:} $\texttt{row}(y,x) \land \texttt{col}(z,y) \Rightarrow \texttt{reduction}(z,x)$
\end{adjustbox}
\begin{adjustbox}{max width=\linewidth}
\textbf{Case $<$:} $\texttt{row}(y,x) \land \texttt{col}(y,z) \Rightarrow \texttt{reduction}(x,z)$
\end{adjustbox}
\end{subfigure}
\captionof{figure}{Reduction Matrix. For each set of two reasoning skills (row then column), there are four permutations of the ordering of the variables $x$, $y$, and $z$. For each case, we manually determine whether a reduction is entailed, and, if so, its skill type (shown here as the first letter of the skill name). For example, Figure \ref{fig:formal-vs-commonsense} illustrates case `$\leftarrow$' of \texttt{spatial} $\land$ \texttt{part\_of}. Since cases are symmetric, only the upper triangle is shown.}
\label{fig:reduction_matrix}
\end{figure}

To automatically generate reasoning trees of arbitrary size, we need to compose reasoning skills. However, not all combinations of reasoning skills form a valid composition. Specifically, we consider a set of two reasoning skills to be compositionally valid only if, together, they consistently \textit{reduce} to another commonsense skill. For example, $\texttt{part\_of}(x,y) \land \texttt{spatial}(y,z)$ reduce to $\texttt{spatial}(x,z)$---if a whole $y$ is near an object $z$, then any part $x$ of that whole $y$ must also be near $z$. One the other hand, $\texttt{spatial}(x,y) \land \texttt{causal}(y,z)$ are not reducible---just because some object $x$ is near a cause $y$ does not mean that $x$ is also near the effect $z$, nor that $x$ is an indirect cause of $z$.

Reducibility thus ensures that templates are composed commonsensically, rather than arbitrarily. However, due to the informal nature of commonsense reasoning, we determined reduction validity based on unanimous consensus among authors, rather than formal proofs. That said, we provide informal proofs below as commonsense rationale for each reduction.

For each set of two reasoning skills, we manually determine its reduction (if any) and keep only those whose reduction is both (a) valid and (b) closed under the set of skills. The results are summarized in Figure \ref{fig:reduction_matrix}. The closure constraint is a practical consideration to speed up the algorithm, as it removes recursive dead ends in which unverifiable reasoning paths are generated. Specifically, closure naturally ensures reasoning paths of any length always have corresponding entries in the reduction matrix lookup table.

One small nuance here is that negation can inadvertently invalidate some reductions. To illustrate this point, recall that in all reductions, one relation type always \textit{dominates} over the other. For the 6 transitive reductions, we assume that both relations can dominate. As a result, although permutations cases are always symmetric in their logic forms, the transitive cases are not necessarily symmetric with respect to their template surface forms. For example, consider $\texttt{causal}(x, y) \land \texttt{causal}(y, z) \Rightarrow \texttt{causal}(x, z)$. If the first relation is the pairing template (and therefore dominates), then the surface form of the second relation becomes ``Suppose that \textbf{only} [$y$] causes [$z$].'' For the inverse, the surface form of the first relation becomes ``Suppose that [$y$] \textbf{only} causes [$z$].'' The placement of ``only'' in these surface forms is critical to ensuring logical soundness when negating the pairing template. For example, ``[$x$] does not cause [$y$]'' and ``[$y$] causes [$z$]'' does \textit{not} imply that ``[$x$] does not cause [$z$],'' since it remain logically possible for $x$ to cause $z$ directly. In contrast, ``[$x$] does not cause [$y$]'' and ``\textbf{only} [$y$] causes [$z$]'' does imply that ``[$x$] does not cause [$z$],'' since the only direct cause of $z$ is $y$, but $x$ does not cause $y$. We track these surface form variations according to which template position in the reduction is matched against the pairing template, which is done independently of the (purely symmetric) reduction matrix.\footnote{In principle, instead of tree duplication (see Figure \ref{fig:csqa:duplication}), each answer choice $a_i$ could have its own pairing template, or at least its own pairing term. Doing so could in principle remove the need for statement negation as a means of answer selection. However, doing so would also make it nearly impossible to programmatically ensure soundness across the entire CSQA instance. In contrast, having all else being equal \textit{except} negation ensures much cleaner mutual exclusivity of the answer choices.}

For each combination of two template types, there are four possible permutation cases of their variables that need validating (see Figure \ref{fig:reduction_matrix}). Given that we have 6 skill types, we have in total $6 \times 6 \times 4 = 144$ validations to consider. However, the permutations cases are symmetric. As such, we only need to consider around half of the cases in practice (specifically, 78, the upper triangle and the main diagonal in the reduction matrix). These are the entries in the matrix in Figure \ref{fig:reduction_matrix}. Of these, we found that 17 cases are valid, including variations under negation. Below, we provide a commonsense rationale for each of these valid cases.

\paragraph{Validated by transitivity:}
\begin{itemize}
\item $\texttt{spatial}(x, y) \land \texttt{spatial}(y, z)$

\hspace{10em}$\Rightarrow \texttt{spatial}(x, z)$

By transitivity of spatial location.
\item $\texttt{causal}(x, y) \land \texttt{causal}(y, z)$

\hspace{10em}$\Rightarrow \texttt{causal}(x, z)$

By transitivity of causality.
\item $\texttt{part\_of}(x, y) \land \texttt{part\_of}(y, z)$

\hspace{10em}$\Rightarrow \texttt{part\_of}(x, z)$

By transitivity of meronymy.
\item $\texttt{type\_of}(x, y) \land \texttt{type\_of}(y, z)$

\hspace{10em}$\Rightarrow \texttt{type\_of}(x, z)$

By transitivity of hypernymy.
\item $\texttt{used\_for}(x, y) \land \texttt{used\_for}(y, z)$

\hspace{10em}$\Rightarrow \texttt{used\_for}(x, z)$

By transitivity of purpose.
\item $\texttt{requires}(x, y) \land \texttt{requires}(y, z)$

\hspace{10em}$\Rightarrow \texttt{requires}(x, z)$

By transitivity of precondition.
\end{itemize}

\paragraph{Validated by hypernymy:}
\begin{itemize}
\item $\texttt{spatial}(x, y) \land \texttt{type\_of}(z, y)$

\hspace{10em}$\Rightarrow \texttt{spatial}(x, z)$

If $x$ appears near all objects of type $y$ and if $z$ is an object of type $y$, then $x$ is near $z$.
\item $\texttt{type\_of}(x, y) \land \texttt{spatial}(y, z)$

\hspace{10em}$\Rightarrow \texttt{spatial}(x, z)$

If all objects of type $y$ appear near object $z$ and if $x$ is an object of type $y$, then $x$ is near $z$.
\item $\texttt{causal}(x, y) \land \texttt{type\_of}(z, y)$

\hspace{10em}$\Rightarrow \texttt{causal}(x, z)$

If $x$ is the cause of all effects of type $y$ and if $z$ is an effect of type $y$, then $x$ causes $z$.
\item $\texttt{type\_of}(x, y) \land \texttt{causal}(y, z)$

\hspace{10em}$\Rightarrow \texttt{causal}(x, z)$

If any cause of type $y$ causes effect $z$ and if $x$ is a cause of type $y$, then $x$ causes $z$.
\item $\texttt{part\_of}(x, y) \land \texttt{type\_of}(z, y)$

\hspace{10em}$\Rightarrow \texttt{part\_of}(x, z)$

If $x$ is a part of all objects of type $y$ and if $z$ is an object of type $y$, then $x$ is a part of $z$.
\item $\texttt{type\_of}(x, y) \land \texttt{used\_for}(y, z)$

\hspace{10em}$\Rightarrow \texttt{used\_for}(x, z)$

If all objects/actions of type $y$ have purpose $z$ and $x$ is an object of type $y$, then $x$ has purpose $z$.
\item $\texttt{type\_of}(x, y) \land \texttt{requires}(y, z)$

\hspace{10em}$\Rightarrow \texttt{requires}(x, z)$

If all objects/actions of type $y$ require $z$ and $x$ is an object of type $y$, then $x$ requires $z$.
\end{itemize}

\paragraph{Others:}
\begin{itemize}
\item $\texttt{spatial}(x, y) \land \texttt{part\_of}(y, z)$

\hspace{10em}$\Rightarrow \texttt{spatial}(x, z)$

Assuming \texttt{part\_of} relates two concrete nouns (which is generally true in ConceptNet), a meronym is typically spatially near its holonym. That is, if $x$ is near $y$ and $y$ is a part of (and therefore near) $z$, then $x$ is near $z$.
\item $\texttt{part\_of}(x, y) \land \texttt{spatial}(y, z)$

\hspace{10em}$\Rightarrow \texttt{spatial}(x, z)$

Assuming \texttt{part\_of} relates two concrete nouns (which is generally true in ConceptNet), a meronym is typically spatially near its holonym. That is, if $x$ is a part of (and therefore near) $y$ and $y$ is near $z$, then $x$ is near $z$.
\item $\texttt{causal}(x, y) \land \texttt{used\_for}(y, z)$

\hspace{10em}$\Rightarrow \texttt{used\_for}(x, z)$

If $x$ causes $y$ whose purpose is $z$, then $x$ is being used to cause $z$. For example, $\texttt{causal}(\texttt{playing piano}, \texttt{noise})$ $\land$ $\texttt{used\_for}(\texttt{noise}, \texttt{distracting others})$ $\Rightarrow$ $\texttt{used\_for}(\texttt{playing piano},$ $ \texttt{distracting others})$. That is, because playing piano causes noise and because noise can be used to distract others, playing piano can be used to distract others.
\item $\texttt{used\_for}(x, y) \land \texttt{requires}(z, y)$

\hspace{10em}$\Rightarrow \texttt{used\_for}(x, z)$

If $y$ is the purpose of $x$ and $z$ requires $y$, then $z$ becomes the purpose of $x$. For example, $\texttt{used\_for}(\texttt{cafe}, \texttt{meeting people})$ $\land$ $\texttt{requires}(\texttt{making friends},$ $\texttt{meeting people})$ $\Rightarrow$ $\texttt{used\_for}(\texttt{cafe}, \texttt{making friends})$. That is, since a cafe can be used for meeting people, and since making friends requires meeting people, then a cafe can be used for making friends.
\end{itemize}

\section{Using a \Factual{} Knowledge Base (ConceptNet) to Ground Variables \orangeemph{Anti-Factually}}
\label{sec:appendix:conceptnet}

To anti-factually ground variables in these reasoning trees, we retrieve assertions from ConceptNet. Specifically, for each duplicated reasoning tree, we treat the pairing term and answer choice as two \textit{seed terms} (see Figure \ref{fig:accord}). From these, we \orangeemph{anti-factually} ground all other variables using a backtracking beam tree search algorithm. Essentially, these seed terms, along with the skill types of the templates in a reasoning tree, place constraints on neighbouring variables. These constraints filter out potential candidates drawn from ConceptNet.

Consider a template with reasoning skill type $r(\mathbf{X}, \mathbf{Y})$. For a given seed term $s$, assume without loss of generality that $s$ is the first term: $r(s, \mathbf{Y})$. We retrieve assertions from ConceptNet table $r$, producing a list of \factual{} assertions of the form $r(x_i, y_i)$. Therefore, excluding any assertions where $x_i = s$ ensures retrieval is purely \antifactual. That is, we set $\mathbf{Y} \leftarrow y_i$ for some $y_i$ such that $r(x_i, y_i)$ is in ConceptNet, but $r(s, y_i)$ is not. Whenever a single candidate variable links to multiple seed terms, we take the set intersection of all retrieved assertions. Finally, we select $k$ terms from these assertions, thereby growing the frontier of grounded variables. To avoid exponential compute times as the tree size grows, we keep only $k=1$ potential candidates in this work. We select these by random sampling. However, using some relevance metric---such as semantic distance between the seed and candidate terms or likelihood of the candidate as computed by some LLM---would be a viable alternative.

We backtrack through unsatisfiable constraints until all variables are grounded, resulting in multiple highly-related trees that, crucially, share the same reasoning structure but have different (\antifactual) grounding. In this way, variables are grounded within appropriate constraints based on skill type, $r$, rather than completely arbitrarily.

Still, the LLMs we benchmark tend to refuse answering anti-factual questions, likely because most have post-hoc constraints to curtail misinformation.\footnote{For example: ``The provided statements are incorrect. Therefore, I cannot provide an answer.''}
To overcome this issue, we introduce a simple instruction prompt that encourages LLMs to accept anti-factual contexts. However, we emphasize that prompt engineering is not the focus of this work. Our instruction prompt is shown in Appendix \ref{sec:appendix:examples}.

Our variable-grounding algorithm does rely somewhat on ConceptNet's recall, since a \factual{} statement omitted from ConceptNet will be treated as \antifactual{} by \accord. However, we can probabilistically hedge against this, so the situation is not as dire as it might initially seem. Specifically, because we randomly chose concept pairs from a large dictionary, it is unlikely the two concepts are related. While it is tricky to give concrete probabilistic bounds, we posit that it is a reasonable assumption. Anecdotally, this is seen in the examples shown in Appendix \ref{sec:appendix:examples}, where \textbf{none} of the concept pairs are \blueemph{factually} related. The added check that the concept pair does not appear in ConceptNet provides additional assurances that it is not \factual, further lowering the chance of a mistake. Ideally, we would instead have a complete KB of commonsense facts to ensure 100\% recall, but no such KB exists. In practice, we must rely on mitigating risks. An alternative would be to manually check every statement using crowd workers, but that would entirely defeat the scalability benefit of \accord, which we posit adds more value than catching the (likely) few mistakes in the dataset.

\section{Replicating Experiments}
\label{sec:appendix:experiments}

The source code repository contains detailed information on the exact commands to run to replicate the experiments. Here we describe the details of our experimental procedures.

\subsection{Manual Preprocessing}
\label{sec:appendix:experiments:preprocessing}

\begin{figure}[t]
\textbf{CSQA ID:} 5e260e1d96187716888cbd968010bb65

\textbf{Question:} Where is the closest place from where you could borrow salt?

\textbf{Answer choices:} A: ocean water\hspace{1em}B: table\hspace{1em}C: shaker\hspace{1em}D: neighbor's house\hspace{1em}E: lake

\textbf{Correct answer:} D: neighbor's house

\textbf{Pairing template:} Suppose that [salt] \{does|does not\} appear near [<insert answer choice>]

~

\textbf{CSQA ID:} 2987db72e66f5fa0015ac64f9b3614ec

\textbf{Question:} What do you do in order to fly in airplane?

\textbf{Answer choices:} A: buy tickets\hspace{1em}B: passenger\hspace{1em}C: read\hspace{1em}D: add gas\hspace{1em}E: run through checklists

\textbf{Correct answer:} A: buy tickets

\textbf{Pairing template:} Suppose that [flying in airplane] \{does|does not\} have prerequisite [<insert answer choice>]

~

\textbf{CSQA ID:} 055817d8d703d3c2802545e3fccdcde3

\textbf{Question:} What do humans do to other humans after death?

\textbf{Answer choices:} A: celebrate\hspace{1em}B: burial\hspace{1em}C: life\hspace{1em}D: rebirth\hspace{1em}E: decomposition

\textbf{Correct answer:} B: burial

\textbf{Pairing template:} Suppose that [<insert answer choice>] \{is|is not\} a type of [death ritual]
\caption{Randomly sampled example pairings from \accord$_\text{CSQA}$.}
\label{fig:example_pairings}
\end{figure}

\accord$_\text{CSQA}$ comprises 6 subsets. Each subset, \accord$_\text{CSQA}^N$, $N \in [0, 5]$, corresponds to all instances in which the context contains precisely $N$ statements per CSQA answer choice. The baseline subset, \accord$_\text{CSQA}^0$, corresponds to the subset of 74 CSQA used to generate the rest.

\accord$_\text{CSQA}^1$ contains base 93 instances. For each instance in \accord$_\text{CSQA}^0$, the authors manually wrote between 1 and 3 pairing templates. Specifically, 56 instances have one pairing template, 17 instances have 2, and 1 instance has 3. From this base, we duplicate each instance such that exactly one copy has the \factual{} answer and at least one copy has an \antifactual{} answer. Since there are exactly 4 alternative answer choices, each base can be duplicated up to 5 times (minimum of 2). In the official \accord$_\text{CSQA}$ release, we duplicate twice, leading to a dataset size of $93 \times 2 = 186$.

A list of all 93 pairing templates can be found in the source code repository. In Figure \ref{fig:example_pairings}, we reproduce 3 randomly chosen pairings as examples. Additional examples can be found on in the source code repository. All pairing templates and all other reasoning path templates contain both positive and negative variations. Distractor templates contain no such variations. These variations are required to negate the logic of a reasoning path to ensure that exactly one of the answer choices is implied, while the others are contradicted. The variations are indicated in curly braces in Figure \ref{fig:example_pairings}, where the first form represents implication and the second form represents contradiction.

\subsection{Automated Generation}
\label{sec:appendix:experiments:generation}

\begin{table*}[t]
\centering
\begin{adjustbox}{max width=\linewidth}
\begin{tabular}{l|rrrrrr|r|rr}
\toprule
& \multicolumn{6}{c}{\textbf{Reasoning Hops}} & & \multicolumn{2}{c}{\textbf{Number of Words}} \\
\textbf{Subset} & 0 & 1 & 2 & 3 & 4 & 5 & \textbf{Total} & \textbf{w/ Prompt} & \textbf{Net} \\
\midrule
\accord$_\text{CSQA}^0$ & 74 & $-$ & $-$ & $-$ & $-$ & $-$ & 74 & 4,700 & 2,110 \\
\accord$_\text{CSQA}^1$ & $-$ & 186 & $-$ & $-$ & $-$ & $-$ & 186 & 28,864 & 14,170 \\
\accord$_\text{CSQA}^2$ & $-$ & 27,218 & 8,364 & $-$ & $-$ & $-$ & 35,582 & 7,428,290 & 4,617,312 \\
\accord$_\text{CSQA}^3$ & $-$ & 4,238 & 6,552 & 11,522 & $-$ & $-$ & 22,312 & 5,833,673 & 4,071,025 \\
\accord$_\text{CSQA}^4$ & $-$ & 7,512 & 6,174 & 12,522 & 27,418 & $-$ & 53,626 & 16,627,058 & 12,390,604 \\
\accord$_\text{CSQA}^5$ & $-$ & 12,344 & 10,686 & 11,534 & 27,966 & 71,204 & 133,734 & 48,451,533 & 37,886,547 \\
\midrule
\textbf{Total} & 74 & 51,498 & 31,776 & 35,578 & 55,384 & 71,204 & \textbf{245,514} & 78,374,118 & 58,981,758 \\
\bottomrule
\end{tabular}
\end{adjustbox}
\caption{Maximum size for each data subset.}
\label{tab:max_totals}
\end{table*}

\begin{table*}[t]
\centering
\begin{adjustbox}{max width=\linewidth}
\begin{tabular}{l|rrrrrr|r|rr}
\toprule
& \multicolumn{6}{c}{\textbf{Reasoning Hops}} & & \multicolumn{2}{c}{\textbf{Number of Words}} \\
\textbf{Subset} & 0 & 1 & 2 & 3 & 4 & 5 & \textbf{Total} & \textbf{w/ Prompt} & \textbf{Net} \\
\midrule
\accord$_\text{CSQA}^0$ & 74 & $-$ & $-$ & $-$ & $-$ & $-$ & 74 & 4,700 & 2,110 \\
\accord$_\text{CSQA}^1$ & $-$ & 186 & $-$ & $-$ & $-$ & $-$ & 186 & 28,864 & 14,170 \\
\accord$_\text{CSQA}^2$ & $-$ & 186 & 186 & $-$ & $-$ & $-$ & 372 & 77,756 & 48,368 \\
\accord$_\text{CSQA}^3$ & $-$ & 186 & 186 & 186 & $-$ & $-$ & 558 & 145,682 & 101,600 \\
\accord$_\text{CSQA}^4$ & $-$ & 186 & 186 & 186 & 186 & $-$ & 744 & 231,834 & 173,058 \\
\accord$_\text{CSQA}^5$ & $-$ & 186 & 186 & 186 & 186 & 186 & 930 & 337,414 & 263,944 \\
\midrule
\textbf{Total} & 74 & 930 & 744 & 558 & 372 & 186 & \textbf{2,864} & 826,250 & 603,240 \\
\bottomrule
\end{tabular}
\end{adjustbox}
\caption{Size for each data subset used in experiments.}
\label{tab:exp_totals}
\end{table*}

Generating \accord$_\text{CSQA}^N$ where $N > 1$ from the preprocessed data using \accord{} involves dozens of additional hyper-parameters. The source code repository contains configuration files with all parameters set to those used to generate \accord$_\text{CSQA}$. Each hyper-parameter is explained in detail in the source code, as well. Most hyper-parameters have obvious defaults, which we used. However, trial and error was used to select appropriate values for sub-sampling probabilities. Essentially, we want to minimize sub-sampling as much as possible while also keeping the wall time and final dataset size reasonable. For example, \accord$_\text{CSQA}^5$ could, in principle, contain hundreds of millions of instances. Instead, we aggressively filter this down to just 133,734 instances (see Table \ref{tab:max_totals}).


Each reasoning tree consists of templates containing two variables each. Variables are grounded via matches against the associated ConceptNet table in such a way that the templates are always grounded \orangeemph{anti-factually}, despite ConceptNet being a purely \factual{} KB (see Appendix \ref{sec:appendix:conceptnet}). Validation is based on the assumption that ConceptNet entries are indeed \factual, which is generally true despite some noise in ConceptNet (see Appendix \ref{sec:appendix:conceptnet}).

For each subset \accord$_\text{CSQA}^N$ where $N > 1$, the minimum number of base instances is $N \times P$, where $P = 93$ is the number of pairings in \accord$_\text{CSQA}^1$. By design, $N$ corresponds both to the problem size and to the number of unique reasoning hops possible within that subset. As with \accord$_\text{CSQA}^1$, we duplicate each base instance 2 times (one \factual{} and one \antifactual). Altogether, \accord$_\text{CSQA}^N$ contains a minimum of $2 \times N \times P \times R$, where $R > 0$ represents the minimum resampling rate. For each $R$, we resample all non-fixed grounding terms in the instance from ConceptNet to generate new instances with the \textit{same} underlying structure but different grounding terms. In the official \accord$_\text{CSQA}$ release, we set $R \ge 10$, so that the minimum dataset size for \accord$_\text{CSQA}^N$ is $2 \times N \times 93 \times 10 = 1860N$. Since $R$ is a minimum, the official datasets are significantly larger, since some pairing combinations occur much more frequently. We note that not all instances need to be used. We provide a large dataset only to cover potential uses cases that require more data. In particular, the experiments in \S\ref{sec:experiments} only use $R = 1$, which is sufficient to achieve the desired narrowness of the estimated standard error. Table \ref{tab:max_totals} shows the maximum size distributions of the official dataset release. Table \ref{tab:exp_totals} shows the size distribution used for our experiments.

\subsection{Benchmarking LLMs}

For each LLM, we used its default hyper-parameters based on its respective configuration files. We used a chat-style interface instead of a text-only interface (e.g., using OpenAI's Chat Completions API rather that its Completions API). All LLMs generate exactly one output sequence per input. For OpenAI models, we used the JSON object response format type with max tokens of 20. For all other models, we prompt the model to output a JSON-formatted answer, but allow for up to 500 tokens to be generated from which we attempt to extract an answer.\footnote{Our answer extraction success rate is well over 99\%. Answer extraction is based on a complex cascade of JSON parsing and matching against labels or answer terms. See the source code repository for the exact sequence.} Instructions to output in JSON format are appended to the primary instruction prompt shown in the main paper. OpenAI models were evaluated using OpenAI's Chat Completions API, whereas all other models are run locally on A40 GPUs on an internal cluster. Gemma 7B and Mistral-7B-Instruct-v0.1 run through the HuggingFace Transformers API on one A40 each. All other local models run through VLLM served using OpenAI's Chat Completions API. Meta-Llama-3-8B-Instruct ran on one A40. Llama-2-13b-chat-hf ran on two A40s. Llama-2-70b-chat-hf, Meta-Llama-3-70B-Instruct, and Mixtral-8x7B-Instruct-v0.1 ran on four A40s each. Mixtral-8x22B-Instruct-v0.1 ran on eight A40s (four over two nodes). Runtimes are fairly consistent, taking about 12 hours per model.

Our benchmarking metric is accuracy with Wald standard error as our variation indicator. A single evaluation run is performed with a fixed random seed ($=314159$) for all LLMs, but each run includes multiple instances for each category and type (see Table \ref{tab:exp_totals}). Standard error is based on the variation over these instances, not over runs.

Results for additional LLMs beyond those in the main body of the paper are shown in Figure \ref{fig:extra_results}.

\section{Human Assessment and Performance on \accord}
\label{sec:appendix:human}

Under a constrained time budget, humans would likely perform quite poorly on \accord, just as humans would likely struggle with first-order logical reasoning under a constrained time budget. However, as is the case with logical reasoning \cite{wu2023reasoning}, we argue that humans have the \textit{competence} to solve \accord{} tasks, but that doing so robustly would necessitate large time budgets. In general, human logical reasoning relies on System II of the Dual Process Theory \cite{tang2023large}. However, humans also show a ``content effect'' bias wherein logical reasoning performance decreases when the semantics of the logic are counterfactual, especially with a constrained time budget \cite{dasgupta2022language}. LLMs are also affected by such decoupled semantics \cite{dasgupta2022language, tang2023large}. Since \accord{} blends counterfactual commonsense with formal logical elements, it is sensible to assume that solving \accord{} also requires System II and suffers from the same content effect bias.

However, we posit that the purpose of LLMs is \textit{not} to precisely model human intelligence \cite{wu2023reasoning}. As such, human performance on \accord{} is mostly irrelevant to our goal of evaluating LLM reasoning performance (besides providing a baseline for comparison). Since each problem in the dataset \textit{can} be unambiguously solved, we should---in the long run---strive to create AI systems that can reason through these problems (regardless of average human performance), as a proxy for their commonsense reasoning ability.

\section{Rejected CSQA Instances}
\label{sec:appendix:noise}

Figure \ref{fig:noisy_csqa} highlights the two CSQA instances that were rejected as part of the preprocessing steps for \accord$_\text{CSQA}$. In both cases, the instances were rejected because the attempted pairing templates do not quite capture the essence of the question. In the first instance, the question is asking about an action done by an agent to an object (someone moving furniture) whereas the pairing template is differentiating between answer choices based on a property of the object (whether furniture does appear near some location or other object). In the second instance, the answer to the question is a conjunction, which renders the wording of the question too convoluted for our rather rigid template format.

\begin{figure}[t]
\textbf{CSQA ID:} 3e536d9253bfac45de83e8ee291ca143

\textbf{Question:} Where might it be hard to get furniture to?

\textbf{Answer choices:} A: apartment\hspace{1em}B: loft\hspace{1em}C: store\hspace{1em}D: rug\hspace{1em}E: stairs

\textbf{Correct answer:} B: loft

\textbf{Attempted pairing template(s):}

\hspace{2em}Suppose that [furniture] \{does not|does\} appear near [<insert answer choice>]

~

\textbf{CSQA ID:} e56c56c3cfe50ba0c787c2bd67255be8

\textbf{Question:} She asked her little boy why, he replied that he didn't know and it was just what?

\textbf{Answer choices:} A: case\hspace{1em}B: reason\hspace{1em}C: how\hspace{1em}D: because\hspace{1em}E: answer

\textbf{Correct answer:} D: because

\textbf{Attempted pairing template(s):}

\hspace{2em}Suppose that [<insert answer choice>] \{is|is not\} a part of [the unknown]

\hspace{2em}Suppose that [<insert answer choice>] \{is|is not\} a type of [little boy behavior]
\caption{CSQA instances rejected because we were unable to craft valid pairing templates.}
\label{fig:noisy_csqa}
\end{figure}

\section{Additional Random Samples from \accord}
\label{sec:appendix:examples}

These examples are drawn from the small variant of \accord$_\text{CSQA}$. We randomly sampled one example per problem size from 0 to 5, shown in increasing size order. For each, we show the full, complete, raw example---including the instruction prompt for the LLMs and the answer prompt. The LLM does not see the \texttt{Instance ID} or the \texttt{Meta-data}, however. These are shown here for human legibility.

\begin{lstlisting}
Instance ID: G50_0_E
Meta-data:
    Reasoning Hops: 0
    Distractors: 0
    Problem Size: 0
    Ground Truth Label: E

Instructions:
Answer the following multiple-choice question.
Provide your answer in JSON format using the
following schema: {"answer": <label>} where
<label> is exactly one of: "A", "B", "C", "D",
or "E". Do not output anything else.
Question:
He was on trial for obstructing justice, during
which he made a questionable comment and was
also found guilty of what?
A: prosecution    B: getting hurt    C: sweat
D: steam    E: committing perjury
Answer:
\end{lstlisting}

\vspace{3em}

\begin{lstlisting}
Instance ID: G68_1_B
Meta-data:
    Reasoning Hops: 1
    Distractors: 0
    Problem Size: 1
    Ground Truth Label: B

Instructions:
You will be provided with statements relating to
a multiple-choice question. The contents of the
statements may disagree with your prior
knowledge of the world. That is ok. Your task is
to provide the most appropriate answer to the
multiple-choice question based on the reasoning
presented in the statements. Provide your answer
in JSON format using the following schema:
{"answer": <label>} where <label> is exactly one
of: "A", "B", "C", "D", or "E". Do not output
anything else.
Statements:
- Suppose that [sitting_quietly] is not a part of [fall asleep]
- Suppose that [sitting_quietly] is a part of [meditate]
- Suppose that [sitting_quietly] is not a part of [reading]
- Suppose that [sitting_quietly] is not a part of [bunk]
- Suppose that [sitting_quietly] is not a part of [think]
Question:
What is someone doing if he or she is sitting
quietly and his or her eyes are moving?
A: reading    B: meditate    C: fall asleep
D: bunk    E: think
Answer:
\end{lstlisting}

\vspace{3em}

\begin{lstlisting}
Instance ID: G16155_2_C
Meta-data:
    Reasoning Hops: 2
    Distractors: 0
    Problem Size: 2
    Ground Truth Label: C

Instructions:
You will be provided with statements relating to
a multiple-choice question. The contents of the
statements may disagree with your prior
knowledge of the world. That is ok. Your task is
to provide the most appropriate answer to the
multiple-choice question based on the reasoning
presented in the statements. Provide your answer
in JSON format using the following schema:
{"answer": <label>} where <label> is exactly one
of: "A", "B", "C", "D", or "E". Do not output
anything else.
Statements:
- Suppose that [serious] is a type of [toilet training product]
- Suppose that [longplay] is a type of [pinniped mammal]
- Suppose that [mammalogy] is a type of [boring activity]
- Suppose that [coccid insect] is a type of [boring activity]
- Suppose that [musical] is a type of [entree]
- Suppose that [pinniped mammal] is a type of [boring activity]
- Suppose that [toilet training product] is a type of [boring activity]
- Suppose that [eat cake] is a type of [coccid insect]
- Suppose that [doing nothing] is a type of [mammalogy]
- Suppose that [entree] is not a type of [boring activity]
Question:
Sarah didn't like to play but she didn't want to
be sedentary and bored, either, so she took up
what?
A: serious    B: longplay    C: musical
D: eat cake    E: doing nothing
Answer:
\end{lstlisting}

\vspace{3em}

\begin{lstlisting}
Instance ID: G7186_3_C
Meta-data:
    Reasoning Hops: 3
    Distractors: 0
    Problem Size: 3
    Ground Truth Label: C

Instructions:
You will be provided with statements relating to
a multiple-choice question. The contents of the
statements may disagree with your prior
knowledge of the world. That is ok. Your task is
to provide the most appropriate answer to the
multiple-choice question based on the reasoning
presented in the statements. Provide your answer
in JSON format using the following schema:
{"answer": <label>} where <label> is exactly one
of: "A", "B", "C", "D", or "E". Do not output
anything else.
Statements:
- Suppose that [each country] appears near [breakfast cereal]
- Suppose that [a steering wheel] does appear near [each country]
- Suppose that [vase] appears near [aetna]
- Suppose that [jumbo jet] appears near [preserved foods]
- Suppose that [parcel] appears near [train]
- Suppose that [display] appears near [parcel]
- Suppose that [a steering wheel] does not appear near [motels]
- Suppose that [a steering wheel] does not appear near [vase]
- Suppose that [preserved foods] appears near [drawer]
- Suppose that [traffic signs] appears near [firearm]
- Suppose that [aetna] appears near [keep cloesd]
- Suppose that [motels] appears near [traffic signs]
- Suppose that [breakfast cereal] appears near [ignition switch]
- Suppose that [a steering wheel] does not appear near [jumbo jet]
- Suppose that [a steering wheel] does not appear near [display]
Question:
The lock kept the steering wheel from moving,
but the thief still took his chances and
began to work on the what?
A: keep cloesd [sic]    B: train    
C: ignition switch D: drawer    E: firearm
Answer:
\end{lstlisting}

\vspace{3em}

\begin{lstlisting}
Instance ID: G13526_4_C
Meta-data:
    Reasoning Hops: 4
    Distractors: 0
    Problem Size: 4
    Ground Truth Label: C

Instructions:
You will be provided with statements relating to
a multiple-choice question. The contents of the
statements may disagree with your prior
knowledge of the world. That is ok. Your task is
to provide the most appropriate answer to the
multiple-choice question based on the reasoning
presented in the statements. Provide your answer
in JSON format using the following schema:
{"answer": <label>} where <label> is exactly one
of: "A", "B", "C", "D", or "E". Do not output
anything else.
Statements:
- Suppose that [kissing too long] does not cause [feet touch ground]
- Suppose that only [change] causes [charge]
- Suppose that only [understanding better] causes [dry pet food]
- Suppose that [strong feelings] is a type of [charge]
- Suppose that [arousal] is a type of [morning glory]
- Suppose that only [working and getting paid for] causes [understanding better]
- Suppose that only [feet touch ground] causes [lying]
- Suppose that [kissing too long] does cause [working and getting paid for]
- Suppose that [kissing too long] does not cause [hurting]
- Suppose that only [returning to work] causes [kingdom]
- Suppose that only [hurting] causes [driving]
- Suppose that [herpes] is a type of [kingdom]
- Suppose that [kissing too long] does not cause [leaving ring]
- Suppose that [excitement] is a type of [character set]
- Suppose that only [driving] causes [character set]
- Suppose that [kissing too long] does not cause [use brain]
- Suppose that [shortness of breath] is a type of [dry pet food]
- Suppose that only [use brain] causes [change]
- Suppose that only [leaving ring] causes [returning to work]
- Suppose that only [lying] causes [morning glory]
Question:
What happens if someone kisses too long?
A: strong feelings    B: herpes    C: shortness of
breath    D: excitement    E: arousal
Answer:
\end{lstlisting}

\vspace{3em}

\begin{lstlisting}
Instance ID: G31713_5_D
Meta-data:
    Reasoning Hops: 1
    Distractors: 4
    Problem Size: 5
    Ground Truth Label: D

Instructions:
You will be provided with statements relating to
a multiple-choice question. The contents of the
statements may disagree with your prior
knowledge of the world. That is ok. Your task is
to provide the most appropriate answer to the
multiple-choice question based on the reasoning
presented in the statements. Provide your answer
in JSON format using the following schema:
{"answer": <label>} where <label> is exactly one
of: "A", "B", "C", "D", or "E". Do not output
anything else.
Statements:
- Suppose that [not losing weight] does not cause [beauty]
- Suppose that [protecting passport] is a type of [catholicism]
- Suppose that [loose skin] is used for [privacy]
- Suppose that [placing basketball] causes [not losing weight]
- Suppose that [privacy] is a type of [restraint]
- Suppose that [healthier] is used for [protecting passport]
- Suppose that [death] is used for [writing to friend or business]
- Suppose that [not losing weight] does not cause [loose skin]
- Suppose that [orient] causes [not losing weight]
- Suppose that [text string occurrence] is used for [achieving goal]
- Suppose that [catholicism] is used for [cook oatmeal]
- Suppose that [depression] causes [not losing weight]
- Suppose that [not losing weight] does not cause [healthier]
- Suppose that [not losing weight] does not cause [miss universe]
- Suppose that [writing to friend or business] is a type of [text string occurrence]
- Suppose that [using water colors] is a type of [vendor]
- Suppose that [invite people over] is a type of [sputnik]
- Suppose that [beauty] is used for [invite people over]
- Suppose that [watering lawn] causes [not losing weight]
- Suppose that [familiar sound] causes [not losing weight]
- Suppose that [not losing weight] does cause [death]
- Suppose that [sputnik] is used for [getting up in morning]
- Suppose that [miss universe] is used for [using water colors]
- Suppose that [vendor] is used for [transporting cargo]
- Suppose that [restraint] is used for [avoid sunburn]
Question:
What might happen if someone is not losing
weight?
A: loose skin    B: beauty    C: miss universe
D: death    E: healthier
Answer:
\end{lstlisting}

\section{Scientific Artifacts}
\label{sec:appendix:assets}

\subsection{Data}

\accord$_\text{CSQA}$ builds from CSQA \cite{talmor2018commonsenseqa} and ConceptNet \cite{speer2017conceptnet}. To the best of our knowledge, CSQA was released without a license. In particular, neither the official homepage\footnote{\url{https://www.tau-nlp.org/commonsenseqa}} nor the GitHub repository\footnote{\url{https://github.com/jonathanherzig/commonsenseqa}} include licensing information as of Oct 14, 2024. ConceptNet data is released under the Creative Commons Attribution Share-Alike 4.0 License.

The \accord$_\text{CSQA}$ data is released under the Creative Commons Attribution Share-Alike 4.0 License. The \accord{} code and the code to generate the \accord$_\text{CSQA}$ data is released under the MIT License.

\subsection{Models}

\accord$_\text{CSQA}$ was benchmarked against various large language models (LLMs). The license for each benchmarked LLM is as follows:
\begin{itemize}
\item Gemma 7B (gemma-7b-it) \cite{team2024gemma}: Gemma Terms of Use \cite{gemmalicense}
\item GPT 3.5 (gpt-3.5-turbo-0125) \cite{gpt35}: OpenAI Terms of Use \cite{gptlicense}
\item GPT-4o (gpt-4o-2024-05-13) \cite{gpt4o}: OpenAI Terms of Use \cite{gptlicense}
\item Llama-2-13b-chat-hf \cite{touvron2023llama}: Llama 2 Community License Agreement \cite{llama2license}
\item Llama-2-70b-chat-hf \cite{touvron2023llama}: Llama 2 Community License Agreement \cite{llama2license}
\item Meta-Llama-3-70B-Instruct \cite{llama3modelcard}: Llama 3 Community License Agreement \cite{llama3license}
\item Meta-Llama-3-8B-Instruct \cite{llama3modelcard}: Llama 3 Community License Agreement \cite{llama3license}
\item Mistral-7B-Instruct-v0.1 \cite{jiang2023mistral}: Apache License 2.0
\item Mixtral-8x22B-Instruct-v0.1 \cite{mixtralmodelcard}: Apache License 2.0
\item Mixtral-8x7B-Instruct-v0.1 \cite{jiang2024mixtral}: Apache License 2.0
\end{itemize}

\section{Societal Impact}
\label{sec:appendix:ethics}

\accord{} was created to work towards closing the measurability gap between commonsense and formal reasoning tasks for LLMs. LLMs can perform remarkably well on reasoning tasks. However, detailed analysis reveals limitations in their reasoning that typically worsens with increased reasoning complexity. Commonsense reasoning is especially challenging for LLMs. As such, key players in AI have singled out commonsense as a critical new frontier. Unfortunately, a detailed understanding of LLMs' commonsense reasoning abilities is severely lagging compared to our understanding of their formal reasoning abilities, since commonsense benchmarks are difficult to construct in a manner that is rigorously quantifiable. Specifically, prior commonsense reasoning benchmarks and datasets are limited to one- or two-hop reasoning or include an unknown (i.e., non-measurable) number of reasoning hops and/or distractors. Arbitrary scalability via compositional construction is also typical of formal reasoning tasks but lacking in commonsense reasoning. Finally, most prior commonsense benchmarks either are limited to a single reasoning skill or do not control skills. Our approach aims to address all these gaps, as discussed in detail in the main body of the paper.

As a standalone benchmark, \accord$_\text{CSQA}$ is unlikely to pose any additional risks or harms beyond those already present in CSQA and ConceptNet (which are themselves very limited due to their genericity). The goal of \accord$_\text{CSQA}$ is to benchmark the commonsense reasoning abilities of LLMs. Unfortunately, depending on the care taken by their creators, LLMs may have significant negative societal impact, including on safety, security, discrimination, surveillance, deception, harassment, human rights, bias, and fairness \cite{bender2021dangers, weidinger2021ethical}. Our aim with \accord$_\text{CSQA}$ is to improve the reasoning abilities of LLMs. Improved reasoning could lead to societal benefits (e.g., LLMs could better reason about the facts of a person, rather than engaging in discriminatory or biased stereotyping). On the hand, a bad actor could leverage this increased reasoning capability---or any increase in LLM capability---to increase societal harm (e.g., more convincing or targeted exploitation or harassment). LLMs also incur significant environmental impacts, in terms of both increased mining of the raw materials needed to produce hardware (e.g., GPUs) and increased electricity consumption needed to power said hardware \cite{bender2021dangers}. In designing \accord{} and \accord$_\text{CSQA}$, we have taken steps to encourage LLMs to genuinely reason about the context provided, while discouraging the reliance on (potentially biased) parametric ``guessing'' of the answer. We have also ensured that \accord$_\text{CSQA}$ can be used as a small or as a large benchmark (see Tables \ref{tab:max_totals} and \ref{tab:exp_totals}), which enables users to chose a size that is only as large as needed for their use case, which can help mitigate some environmental impact.

In addition, \citet{davis2023benchmarks} argues that genuine improvements in LLMs can only be quantified by high-quality benchmarks that focus on a foundational commonsense understanding (including ``basic temporal, spatial, physical, psychological, and social reasoning'') and extend the scope of reasoning complexity beyond the typical one or two hops. In terms of foundational commonsense understanding, we include a broad set of reasoning skills precisely for this reason, and our reasoning trees are explicit attempts at increasing the reasoning complexity compared to prior commonsense reasoning benchmarks. As for quality, although our quality is somewhat limited by the quality of CSQA, we have taken steps to ensure \accord$_\text{CSQA}$ is as high quality as possible. First, our pairing templates are handwritten to counteract any quality concerns from CSQA as much as possible. In particular, they are written to clearly differentiate the answer choices. Our experiments showed that this effort resulted in LLMs improving in performance compared to base CSQA when given the pairing template as a single reasoning hop. Second, our reduction matrix is validated manually to ensure correctness. Third, our \antifactual{} variables are grounded from samples randomly drawn only from the ConceptNet table associated with each template, rather than arbitrarily.

\begin{figure*}[ht]
\centering
\begin{subfigure}{0.5425\linewidth}
\includegraphics[width=\linewidth]{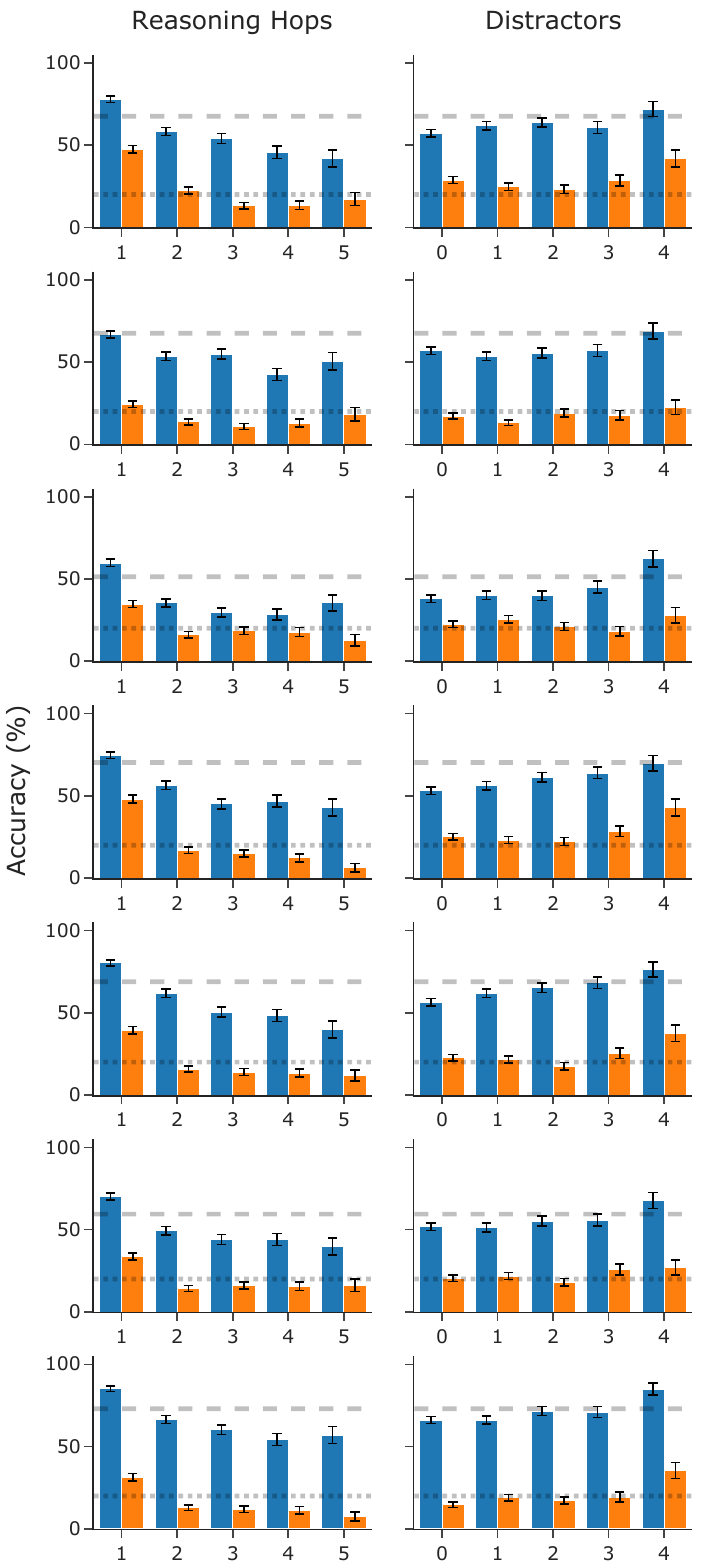}
\end{subfigure}
\begin{subfigure}{0.451\linewidth}
\includegraphics[width=\linewidth]{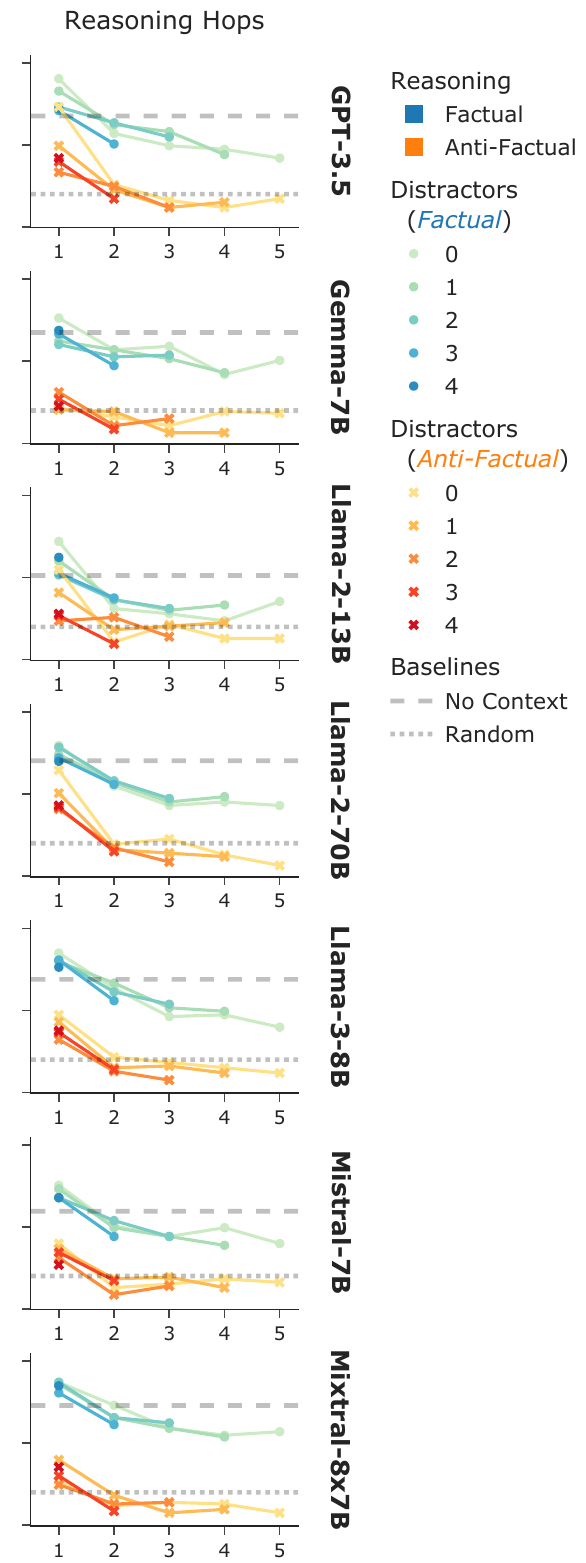}
\end{subfigure}
\caption{Performance of additional LLMs on \accord$_\text{CSQA}$. \textbf{Left:} Both \factual{} and \antifactual{} performance degrade rapidly with increasing reasoning hops, which is expected. \textbf{Middle:} Both \factual{} and \antifactual{} performance increase with increasing distractors, which is unexpected. \textbf{Right:} Disentangling the interaction effect between reasoning hops and distractors. Reasoning hops are dominant while distractors' effect is negligible, which explains the reversed trend in \textbf{(Middle)} that marginalizes over reasoning hops. \textbf{All:} \factual{} significantly outperforms \antifactual, which indicates context unfaithfulness. \Antifactual{} performance drops below random chance as a result. Wald standard error bars are with respect to the 93 pairings, not reruns based on random seeds.}
\label{fig:extra_results}
\end{figure*}

\end{document}